\RequirePackage{booktabs}
\documentclass[pdflatex,sn-vancouver-ay]{sn-jnl}

\usepackage{graphicx}%
\usepackage{multirow}%
\usepackage{amsmath,amssymb,amsfonts}%
\usepackage{amsthm}%
\usepackage{mathrsfs}%
\usepackage[title]{appendix}%
\usepackage{xcolor}%
\usepackage{textcomp}%
\usepackage{manyfoot}%
\usepackage{booktabs}%
\usepackage{algorithmic}
\usepackage{algorithm}
\usepackage{listings}%
\usepackage{xcolor}
\usepackage{amsmath}
\usepackage{graphicx}
 \usepackage{amsfonts}
 \usepackage{multirow}

\usepackage[table]{xcolor}

\usepackage{amssymb}
\usepackage{subfig}
\usepackage{afterpage}
\usepackage[table]{xcolor}

\usepackage{amsmath}
\usepackage{multirow}
\usepackage[graphicx]{realboxes}

\theoremstyle{thmstyleone}%
\theoremstyle{thmstyletwo}%

\theoremstyle{thmstylethree}%

\begin{document}

\title[Active Learning for Anomaly Detection]{Ranking-Enhanced Anomaly Detection Using Active Learning-Assisted Attention Adversarial Dual AutoEncoders}


\author*[1]{\fnm{Sidahmed} \sur{Benabderrahmane}}\email{sidahmed.benabderrahmane@nyu.edu}

\author[2]{\fnm{James} \sur{Cheney}}

\author[1]{\fnm{Talal} \sur{Rahwan}}

\affil*[1]{\orgname{New York University}, \orgdiv{NYUAD, Division of Science}}

\affil[2]{ \orgname{The University of Edinburgh}, \orgdiv{School of Informatics}, \orgaddress{ \country{UK}}}


\abstract{Advanced Persistent Threats (APTs) pose a significant challenge in cybersecurity due to their stealthy and long-term nature. Modern supervised learning methods require extensive labeled data, which is often scarce in real-world cybersecurity environments. In this paper, we propose an innovative approach that leverages AutoEncoders for unsupervised anomaly detection, augmented by active learning to iteratively improve the detection of APT anomalies. By selectively querying an oracle for labels on uncertain or ambiguous samples, we minimize labeling costs while improving detection rates, enabling the model to improve its detection accuracy with minimal data while reducing the need for extensive manual labeling. We provide a detailed formulation of the proposed Attention Adversarial Dual AutoEncoder-based anomaly detection framework and show how the active learning loop iteratively enhances the model. The framework is evaluated on real-world imbalanced provenance trace databases produced by the DARPA Transparent Computing program, where APT-like attacks constitute as little as 0.004\% of the data. The datasets span multiple operating systems, including Android, Linux, BSD, and Windows, and cover two attack scenarios. The results have shown significant improvements in detection rates during active learning and better performance compared to other existing approaches. \\
%
}

\keywords{Anomaly Detection, Deep Learning, Attention Mechanism, AutoEncoders, Active Learning, Generative Adversarial Neural Networks, Cyber-security, Advanced Persistent Threats.}



\maketitle

\section{Introduction}
In today's digital age, the prevalence of cyberattacks has increased drastically, putting critical infrastructure, government agencies, corporations, and private data at risk \cite{zhang2025llms,qawasmeh2025navigating,sood2012}. Among the most sophisticated and dangerous forms of cyberattacks are \textit{Advanced Persistent Threats} (APTs), which are characterized by stealth, persistence, and a high degree of technical sophistication \cite{sharma2023advanced}. APTs often infiltrate networks, remaining undetected for extended periods, with the aim of stealing sensitive data or sabotaging operations \cite{SALIM2023e17156}. Due to their long-term and tailored nature, detecting APTs poses a significant challenge to traditional cybersecurity tools, such as antivirus software or firewalls, which primarily rely on signature-based detection \cite{buchta2024advanced}.

APT actors use a combination of social engineering, zero-day vulnerabilities, custom malware, and lateral movement techniques to infiltrate a system and remain hidden while maintaining access to the network. This advanced nature makes them particularly difficult to detect with traditional signature-based security tools, which rely on predefined patterns of known malware or exploits. Since APTs often blend into normal network behavior and use legitimate credentials or sophisticated obfuscation techniques, they rarely trigger alarms in conventional systems \cite{buchta2024advanced,tankard2011advanced}.

Safeguarding information systems from APT attacks remains a complex challenge. Numerous efforts have been made to design cybersecurity and intelligence tools, with Intrusion Detection Systems (IDS) being a key method to help organizations bolster their defenses against APTs \cite{ussath2016advanced}. Some solutions advocate for the use of network firewalls, which are security devices that monitor Internet traffic and filter unwanted content \cite{kshirsagar2023towards}. Other organizations opt for manual threat hunting strategies that involve security experts who continuously monitor networks around the clock to detect potential threats. In addition, partnering with cybersecurity service providers is another option; in the event of an attack, these companies can provide crucial assistance in responding to threats.

Although these IDS solutions offer considerable promise for detecting Advanced Persistent Threats (APTs), they are hindered by significant limitations related to the substantial investments required in time, personnel, and hardware resources. In cybersecurity, rapid response is paramount, as defensive measures must be deployed more swiftly than attackers can execute their intrusions. The concept of "breakout time", referring to the duration between an intruder gaining initial access and beginning to compromise the network, underscores the necessity of a fast, automated defense system capable of replicating human-like intelligence for effective threat mitigation \cite{alshamrani2019survey}.

Recently, anomaly detection (AD) has emerged as one of the key strategies in identifying cyber-security threats by detecting deviations from normal behavior. Unlike traditional IDS mechanisms, which rely on predefined signatures or rules, anomaly detection works by modeling typical behavior and detecting deviations that may indicate a breach.

However, since APT attacks often involve previously unseen exploits or behaviors not present in signature databases, therefore anomaly detection in the context of APTs presents several challenges, such as:
(i) Low signal-to-noise ratio: In large networks, legitimate anomalies—for instance software updates, high-volume transactions, or administrative activities—can generate significant noise. This makes it difficult for an anomaly detection system to distinguish between benign deviations and true signs of a persistent threat.
(ii) Stealth and low-activity operations: APT attackers operate in a highly stealthy manner, often mimicking legitimate user behavior. They make gradual and subtle changes to the network that are difficult to detect using traditional anomaly detection techniques that focus on detecting more dramatic deviations.
(iii) Label rarity: Most cybersecurity datasets are heavily imbalanced, with a vast majority of data representing normal behavior and only a very small fraction representing anomalous or malicious activity. Moreover, labeled data is scarce, as most attacks are identified post-facto after significant damage has been done, and ground-truth data on APTs is difficult to obtain \cite{buchta2024advanced,cole2012advanced}.

To address these issues and challenges, this paper presents \textit{ALADAEN}, an APT detection framework that comes up with a smart contribution, enabling the model to quickly improve its detection rate while reducing the need for extensive manual labeling and large training datasets, through the combination of Active Learning (\textit{AL}), Generative Adversarial Neural Networks (\textit{GAN}) and an Attention-based Adversarial Dual AutoEncoder (\textit{ADAEN}).  

Firstly, the proposed ADAEN neural network is trained to identify APT-like patterns within imbalanced trace databases, where malicious attacks represent only a small fraction of system activities. This deep neural network learns a low-dimensional representation of normal activity data, and any data that significantly deviates from this representation during decoding is flagged as anomalous and marked for further investigation. The incorporation of the attention mechanism further enhances the model's ability to focus on the most relevant aspects of the input data. Attention mechanisms dynamically assign weights to different parts of the input, allowing the model to prioritize important features in long-sequence of events. The Adversarial Dual AutoEncoder offers, particularly in tasks like anomaly detection, several advantages over a baseline AutoEncoder such as: rich data representation, improved robustness to noise and adversarial attacks, enhanced feature learning, and most importantly support for active learning integration. In fact, the ADAEN neural network is enhanced here with an Active Learning (AL) strategy, that iteratively improves its performance \cite{hsu2015active}. By querying an oracle to label uncertain or anomalous samples, active learning enables a more efficient training process, reducing the amount of labeled data required while improving the model's ability to detect novel and sophisticated attack patterns.

It is worth noting that active learning plays a pivotal role in our framework for APTs detection, especially since there is a scarcity of labeled data. In many real-world scenarios, obtaining a large and well-labeled dataset can be time-consuming, expensive, or impractical, particularly when the data involves rare events such as anomalies or cyberattacks. Our proposed solution addresses this challenge by allowing the model to intelligently select the most uncertain, ambiguous, or potentially informative data points and request labels for only those instances.
Moreover, to further enhance the performance of the model, we incorporate data augmentation using a Generative Adversarial Network (GAN) during active learning \cite{creswell2018generative}. This augmentation helps generate synthetic data resembling the labeled normal points, thereby enriching the training set and improving the model's ability to generalize quickly.\\
The selective querying process makes an efficient use of the limited labeling resources by focusing on the data that will most significantly improve the model performance. In anomaly detection (e.g. APTs), where the dataset is often heavily imbalanced — with very few anomalies compared to normal data — the active learning implementation ensures that the model prioritizes the identification of outliers or unusual patterns. Consequently, this would accelerate the learning process, allowing the model to detect anomalies with higher accuracy even when only a small subset of the data is labeled. By continuously refining the ADAEN neural network with real minimal labeled examples that are augmented with a GAN, active learning minimizes the need for extensive human intervention while significantly enhancing the model's ability to generalize from limited data, ultimately improving detection rates for rare but critical anomalies.

%
%
Our method has been implemented and tested using real provenance data from several operating systems and compared to many existing anomaly detection approaches. It has shown better detection rates than those methods when using the entire training datasets. Furthermore, when using smaller learning datasets, the combination of active learning and GAN allows our approach to iteratively improve its detection rate by selectively augmenting and refining the model's performance over time.

The remainder of the paper is organized as follows.
Section II provides a deeper introduction and discussion on APT attacks, and the anomaly detection methodologies and their application in cyber-security. Section III presents \textit{ALADAEN}, our proposed Active Learning-assisted Attention Adversarial Dual AutoEncoder model for anomaly detection in the context of Advanced Persistent Threat (APT). The section also explains how the framework can be leveraged with the integration of active learning and GANs to refine the detection rates. Section IV summarizes the used datasets, the evaluation metrics, and the experimentation results. Section V concludes the paper with the main outcomes and finally Section VI gives some future perspectives. 
\section{Background and related work}
\subsection{Concept of Advanced Persistent Threats (APTs):}
An Advanced Persistent Threat (APT) is a highly targeted, sophisticated cyberattack in which an attacker gains unauthorized access to a network and remains undetected for an extended period \cite{xuan2024novel}. The primary objectives of APTs are to steal sensitive information, sabotage operations, or establish long-term access to the target network \cite{swetha2025machine}. Unlike typical cyberattacks that are opportunistic and often one-off, APTs are strategic, involving multiple phases over long periods \cite{ghafir2014advanced,shackelford2016protecting}.

APTs typically operate in five distinct phases:
\begin{itemize}
    \item Reconnaissance: The attacker conducts extensive research on the target, identifying vulnerabilities, network architecture, and potential entry points.
 \item Initial Exploitation: The attacker gains access to the target network, often through phishing, exploiting software vulnerabilities, or using stolen credentials.
 \item Establishing Persistence: Once inside, the attacker installs malware, such as back-doors or Remote Access Trojans (RATs), to maintain access to the network even if the initial vulnerability is patched.
 \item Lateral Movement: The attacker escalates privileges and moves laterally through the network to compromise more devices and access sensitive information.
 \item Data Exfiltration/Impact: After gathering valuable information, the attacker exfiltrates the data without raising alarms or uses the compromised systems for further exploitation.
\end{itemize}
Several notable APT attacks over the years have demonstrated the severity and complexity of these threats \cite{dau2024comprehensive,alshamrani2019survey}. They have had far-reaching consequences, targeting critical infrastructure, government organizations, and global corporations \cite{chen2014study}.
Among the most famous APT attacks we can cite: \textit{Goblin Panda APT27}, \textit{Fancy Bear APT28}, \textit{Cozy Bear APT29}, \textit{Ocean Buffalo APT32}, \textit{Helix Kitten APT34}, \textit{Wicked Panda APT41} \textit{APT19}, \textit{Stuxnet}, \textit{Deep panda}, \textit{Epic Turla}, \textit{Oldsmar}, or more recently \textit{Pegasus} \cite{Stuxnet11,marczak2018hide}. This last cyber warfare tool has been capable of reading text messages, tracking calls, collecting passwords, location
tracking, accessing the target device’s microphone and camera,
through a zero-click exploit~\cite{saad2020attribution}. 
These well-known APT attacks illustrate the significant risks posed by APTs and highlight the need for robust detection strategies \cite{han2018tapp,jenkinson2017applying}. Detecting such anomalies requires advanced machine learning techniques to identify subtle deviations in normal patterns \cite{BerradaCBMMTW20}. Indeed, anomaly detection methods, enhanced by modern machine learning techniques, are increasingly being explored as tools for detecting APT activities before they can cause significant harm \cite{Benabderrahmane21}. 
%

Anomaly detection plays a critical role in cybersecurity, as identifying abnormal behaviors within networks or systems is essential for detecting attacks like APTs, zero-day exploits, and insider threats. Over the years, various techniques, ranging from statistical approaches to advanced machine learning models, have been developed to tackle these cyber threats \cite{salim2023systematic,che2024systematic}. Below is an overview of the key methods and recent advancements in their field.
\subsection{Anomaly Detection Methods in Cybersecurity}
\subsubsection{Statistical Methods}  
\begin{itemize}
    \item \textit{Rule-Based Systems:} Early anomaly detection systems relied on predefined rules and thresholds to detect anomalies \cite{ashraf2024hybrid}. For example, intrusion detection systems (IDS) like Snort\footnote{https://www.snort.org/} use signature-based and anomaly-based approaches where deviations from normal patterns trigger alerts. While effective for well-known patterns, these systems struggle to detect novel or evolving threats and can generate a high number of false positives \cite{roesch1999snort}.
    \item \textit{Probability-Based Models:} Techniques such as Gaussian Mixture Models (GMM) and statistical hypothesis testing use probability distributions to model normal system behavior \cite{mangaiyarkaras2025probabilistic,reynoldsg09}. When new data deviates significantly from the modeled behavior, it is flagged as anomalous. Though simple, these models are often limited in capturing the complexity of modern network traffic and application behavior \cite{chandola2009anomaly}.
\end{itemize}

\subsubsection{Machine Learning-Based Approaches}
\begin{itemize}
    \item \textit{Clustering Algorithms}: Unsupervised machine learning models like k-means and DBSCAN have been applied to group data points into clusters, where outliers (i.e., data points that don’t fit any cluster well) are considered anomalous \cite{jain2010data}, \cite{ester1996density},\cite{DBLP:journals/corr/abs-2006-07916}. While clustering can effectively detect outliers, it often struggles with high-dimensional data or imbalanced datasets, which are common in cybersecurity \cite{lazarevic2003comparative}.
   \item \textit{Support Vector Machines (SVMs)}: Support Vector Machines, particularly in their one-class variant (OC-SVM), have been used to define a boundary around normal data in high-dimensional space \cite{scholkopf2001estimating}. Anything falling outside this boundary is classified as an anomaly. Although OC-SVMs perform well for some types of anomalies, they often require extensive tuning and do not scale well for large-scale cybersecurity datasets \cite{algarni2024edge}.
   
   \item \textit{Random Forests and Isolation Forests}: Tree-based methods such as Isolation Forests (iForest) have gained popularity due to their ability to handle high-dimensional data efficiently \cite{liu2008isolation}. Isolation Forests work by isolating anomalies through recursive partitioning of data \cite{benabderrahmane2017combining}. These models are computationally efficient and effective in detecting anomalies without requiring labeled data \cite{nalini2025enhancing}.
\end{itemize}

\subsubsection{Deep Learning Approaches}
\begin{itemize}
    \item  \textit{Recurrent Neural Networks (RNNs) and Long Short-Term Memory Networks (LSTMs)}: Given the sequential nature of network traffic and system logs, RNNs and LSTMs are particularly well-suited for capturing temporal dependencies \cite{bamber2025hybrid}. These models can learn patterns over time and identify deviations in data, making them effective for detecting real-time threats. LSTMs have been used in anomaly detection for detecting anomalies in log sequences, such as unusual access patterns or abnormal process behaviors \cite{malhotra2015lstm,benabderrahmane2017deep}.
    \item Masked Features: Recent advances in self-supervised learning have popularized Masked Autoencoder (MAE) architectures for anomaly detection across diverse domains, including system provenance graphs, network traffic, and IoT telemetry. The core idea of MAE-AD is to randomly mask portions of the input data—whether graph nodes, network flow tokens, or feature vectors—and train the model to reconstruct the missing parts. This mask-and-reconstruct strategy forces the model to learn robust contextual representations of normal behavior without requiring labeled anomalies. In the context of APT detection, MAGIC \cite{jia2024magic} extends this principle to provenance graphs through masked graph representation learning, enabling the capture of rich structural and behavioral dependencies among system entities while remaining fully self-supervised. Similar frameworks, such as the MTC-MAE for malware traffic classification \cite{xu2024self} and CompMAE for IoT intrusion detection \cite{lightbody2025compression}, demonstrate the generality of MAE-AD: they pretrain on large unlabeled datasets to learn unbiased feature embeddings, then fine-tune or apply reconstruction-error-based scoring to flag anomalies. 
    \item Contrastive Learning: To address the limitations of traditional one-class anomaly detectors that rely solely on reconstruction error or distance from a fixed hypersphere, Deep One-Class Contrastive Learning (Deep OCCL) integrates contrastive representation learning into the one-class framework \cite{wang2023deep}. Instead of explicitly modeling normal data boundaries in input space, Deep OCCL learns a compact and discriminative latent manifold by generating multiple augmented views of each normal instance and optimizing a self-supervised contrastive objective that pulls positive pairs together while pushing other samples apart. This dual mechanism enables the network to learn intra-class compactness and inter-instance separability without requiring labeled anomalies, significantly enhancing generalization to unseen attacks or perturbations. During inference, anomalies are identified as samples that lie far from the learned latent center or violate the compact representation structure. 
\end{itemize}

\subsubsection{Hybrid and Ensemble Methods}
\begin{itemize}
    \item \textit{Hybrid Approaches}: Combining multiple models, such as clustering algorithms or SVMs for anomaly classification, can leverage the strengths of different methods \cite{zhang2021federated}. For instance, in hybrid intrusion detection systems (IDS), traditional signature-based methods can be combined with anomaly detection techniques to detect both known and unknown threats \cite{ribeiro2016should}.
   \item \textit{Ensemble Learning}: Techniques like bagging and boosting, where multiple models are trained and their outputs aggregated, are gaining traction in cybersecurity. Isolation Forests, for instance, can be enhanced with ensemble methods to improve accuracy and robustness for attacks detection \cite{bolzoni2014hybrid,dong2020survey}.
   \item Hierarchical Architectures: Among recent developments, High Discrimination APT Intrusion Detection System (HDAPT-IDS) \cite{lee2025ml} represents a hybrid intrusion detection framework that integrates traditional machine learning and deep learning for precise multi-level APT classification. The system is composed of two coordinated modules: the Cyber Clustering Module (CCM) and the Clustering Analysis Module (CAM). The CCM performs a preliminary, coarse-grained categorization of network traffic using a Random Forest model to predict a main class label. Based on this prediction, the CAM dynamically selects a Deep Neural Network specialized for that main class to perform fine-grained sub-class discrimination. This hierarchical design allows HDAPT-IDS to reduce category complexity while improving detection precision. \cite{weng2025rt} proposed RT-APT, a method that constructs provenance graphs from kernel logs and applies the Weisfeiler–Lehman (WL) subtree kernel to capture contextual and structural dependencies between entities. RT-APT leverages the FlexSketch algorithm, which converts provenance graphs into compact sequences of feature vectors that preserve temporal and topological information. Unsupervised K-means clustering is then performed on benign feature vector sequences to model normal system states, enabling the identification of abnormal behaviors that deviate from learned patterns. \cite{liu2025pgae} introduced PGAE, a perturbation-based graph autoencoder framework designed to enhance the robustness and representational power of provenance-based APT detection. The method addresses two major challenges in existing graph neural network approaches: the presence of redundant or noisy edges in provenance graphs, and the limited ability of traditional GNNs to capture low-order semantic feature interactions. To overcome these issues, PGAE employs a dual-edge-node masking mechanism that perturbs the graph structure to improve generalization and robustness. It then utilizes a dual-encoder architecture to jointly learn explicit structural relationships and implicit feature dependencies, followed by a cross-correlation decoder that simultaneously optimizes feature reconstruction and topological rebuilding. This design enables unsupervised node-level APT detection without requiring labeled data. 
\end{itemize}
\subsection{Limitations of Existing Detection Methods}
The use of anomaly detection methods for identifying APTs presents several significant challenges. 
Traditional Intrusion Detection Systems (IDSs) rely on predefined signatures of known threats, 
which are ineffective in identifying new, unknown, or evolving APT tactics that do not match these signatures. 
Even machine-learning-based anomaly detection approaches, while more flexible, tend to generate a high number of false positives, 
mistaking benign activities for potential threats \cite{buchta2024advanced}. 
This can lead to alert fatigue, overwhelming cybersecurity teams and reducing the efficiency of the Security Operations Center (SOC). 
Early-stage APT activities, such as initial compromise or foothold establishment, are especially difficult to detect, 
as they involve subtle actions that blend in with normal system behavior. 
Moreover, APTs frequently employ advanced evasion techniques — such as encryption, polymorphism, and "living-off-the-land" tactics 
that leverage legitimate system tools — making it difficult for conventional IDSs to recognize their presence. 
Additionally, conventional approaches often fail to correlate seemingly unrelated events across multiple hosts and data streams 
that, when combined, are strong indicators of an APT in progress \cite{SALIM2023e17156}.

Beyond signature-based methods, recent deep learning approaches have introduced adversarially trained AutoEncoders 
\cite{zenati2018efficient,akcay2018ganomaly,sabokrou2018alocc} and 
transformer-enhanced variational models \cite{yao2025tii,yin25} for unsupervised anomaly detection. 
While promising, these methods typically assume access to large volumes of clean normal data 
and are trained in a single-shot manner, making them vulnerable to distribution drift and ill-suited to settings 
with extreme label scarcity. 
Similarly, active learning approaches in anomaly detection \cite{audibert2020usad} have largely focused on improving 
classification accuracy rather than optimizing for ranking quality, which is critical for APT triage 
where only a handful of top-ranked alerts can be investigated by SOC analysts. 
Finally, most existing frameworks treat adversarial reconstruction, active querying, and data augmentation as independent modules, 
and do not provide a unified approach to exploit their synergy.

Recent advances outside the cybersecurity domain have proposed sophisticated architectures 
that combine adversarial training and representation learning to address label scarcity. 
USAD employs dual AutoEncoders with adversarial objectives for multivariate time series \cite{audibert2020usad}, 
while Yao et al.\ introduced transformer-enhanced VAEs and scalable Trio-Attention U-Transformers with Siamese discriminators 
for unlabeled anomaly detection in hierarchical IoT settings \cite{yao2025tii,yao2025tsc,yao2025tase}. 
More recently, they proposed a meta-learning powered dual-source representation differentiation framework 
that improves discrepancy analysis for unsupervised anomaly detection. 
Although these works represent significant progress, they primarily target low-dimensional time series or IoT data, 
assume relatively clean training sets, and do not optimize for ranking-centric evaluation. 
Furthermore, they do not offer a unified pipeline combining adversarial learning, data augmentation, and active querying 
in the context of high-dimensional, multi-view provenance graphs.

These limitations motivate the design of {ALADAEN}, which integrates 
(i) dual adversarial AutoEncoders for robust representation learning under extreme imbalance, 
(ii) GAN-based augmentation to densify the normal manifold, and 
(iii) a ranking-oriented active learning loop, 
ensuring that the most informative anomalies are prioritized for SOC analyst feedback.

\subsection{Limitations of Open-source Implementation}
Yet a major obstacle in the field of APT detection is the limited availability of open-source implementations for various proposed methods. This lack of publicly accessible code makes it challenging for researchers to replicate experiments, validate findings, or benchmark new approaches against existing ones. Without a standardized framework or reference implementations, conducting comprehensive comparisons between different APT detection techniques becomes difficult, hindering progress in evaluating their relative strengths and weaknesses across diverse datasets and operational environments. As a result, the development of more effective and universally applicable solutions remains a significant challenge.

\subsection{Limitations of Existing Learning Datasets}

Another major challenge when dealing with APT detection concerns the learning datasets. Broadly, there are several characteristics used to categorize the learning databases in cyber-security, such as  data provenance type, network traffic, system logs, the presence and relevance of labeled attack. Such a categorization would help to better understand the strengths and limitations of each dataset, facilitating the selection of the most suitable data for specific studies. However, there is a noticeable lack of publicly available datasets that accurately capture the behavior of APT attacks. This is a major scientific bottleneck, which significantly hampers the development of effective APT detection models \cite{khraisat2019survey}. \\
Many existing datasets, like KDDCUP98\footnote{http://kdd.ics.uci.edu/databases/kddcup99/kddcup99.html}, NSL-KDD 2009\footnote{https://www.kaggle.com/datasets/hassan06/nslkdd}, UNSW-NB15\footnote{https://research.unsw.edu.au/projects/unsw-nb15-dataset} or CICIDS2017\footnote{https://www.unb.ca/cic/datasets/ids-2017.html}, were designed for general network intrusion detection but do not specifically target APTs \cite{stiawan2020cicids, amine2024optimization}. While many researchers claim to focus on detecting Advanced Persistent Threats (APTs), a large portion of these studies rely on standard intrusion detection system (IDS) datasets that lack APT-specific attack patterns. These datasets typically include common threats such as denial-of-service attacks or basic malware, which do not reflect the complexity, persistence, or stealth of real-world APTs. Consequently, the results of such studies may fall short in addressing the true challenges of APT detection. This raises critical concerns about the relevance and reliability of these works, which often overlook the need for specialized datasets tailored to the unique characteristics of APTs. For example, the work in~\cite{Sakthivelu23} applies ensemble learning to intrusion detection using the \textit{TRAbID} dataset\footnote{https://secplab.ppgia.pucpr.br/?q=trabid}, which focuses exclusively on Remote Desktop Protocol (RDP) netflow events within a single operating system—Windows. As a result, its scope for capturing diverse anomalous behaviors is inherently limited. In~\cite{MARTINLIRAS2021102202}, the authors evaluate several classifiers—including SVM, random forest, kNN, and regression models—for APT detection. However, the dataset consists of Windows log files featuring PE32 executables, object code, and DLLs, based on a legacy 32-bit architecture that no longer reflects current 64-bit systems, thus limiting its relevance. Similarly, the model proposed in~\cite{abdullayeva2021advanced} introduces a deep neural network with additional layers for APT detection, but it was validated solely on a small, Windows-specific malware dataset\footnote{https://marcoramilli.com/2016/12/16/malware-training-sets-a-machine-learning-dataset-for-everyone/} hosted on a personal blog. The dataset lacks documentation on its provenance and curation, and its low citation count suggests limited use in the broader research community.\\
The study in~\cite{app12136816} introduces another deep learning-based APT detection method, trained on a hybrid dataset combining CICIDS2017 and Contagio. While CICIDS2017 includes real-world network traffic, it does not contain APT-like patterns, and although Contagio is claimed to provide APT-relevant data, it is no longer accessible and was also hosted on a personal blog. As such, the validity of the evaluation results cannot be independently verified. Likewise, the authors of~\cite{9496635} validate their method using CICIDS2017 and KDDCUP99—neither of which includes APT attacks. Notably, over half of intrusion detection studies still rely on KDD datasets due to their availability~\cite{AHMED201619net}, despite longstanding criticisms regarding the artificial nature of their generation process~\cite{mchugh2000testing}.

Recent work by~\cite{BerradaCBMMTW20} and~\cite{Benabderrahmane21} has advanced APT detection through system-wide provenance analysis and trace mining. These studies highlight that enriched provenance data can uncover causal dependencies between system events, enabling the detection of sophisticated attack patterns—such as data exfiltration—that often bypass traditional perimeter-based defenses. Their evaluation leveraged the DARPA \verb|ADAPT| dataset, produced under the \verb|Transparent Computing TC| program\footnote{\url{https://www.darpa.mil/program/transparent-computing}}, which offers high-quality, heterogeneous provenance traces. Among the proposed methods, VF-ARM (Valid Frequent Association Rule Mining) flags anomalies when frequent behavioral patterns are violated, while VR-ARM (Valid Rare Association Rule Mining) detects infrequent but valid patterns indicative of anomalous activity~\cite{Benabderrahmane21}. Additional baseline methods include Attribute Value Frequency (AVF)~\cite{koufakou_2007}, which detects rare attribute values, and Frequent Pattern Outlier Factor (FPOF)~\cite{he_2005}, which identifies low-frequency patterns as outliers. Outlier Degree (OD)~\cite{narita_2008} scores anomalies based on association-rule violations, and One-Class Classification by Compression (OC3)~\cite{smets2011} identifies anomalous instances based on reduced compressibility, assuming that anomalies are less redundant than normal patterns.

In this paper, our proposed framework is trained and evaluated using provenance traces from the ADAPT project~\cite{BerradaCBMMTW20}, which includes APT-like attacks across multiple operating systems, including Windows, BSD, Linux, and Android. To our knowledge, the ADAPT project\footnote{\url{https://gitlab.com/adaptdata}} remains the only publicly available source that provides extensive, multi-OS log-event patterns and traces for realistic APT detection research. A significant challenge in training our model with these datasets is that APT attacks make up just 0.004\% of the total samples, creating an extremely imbalanced classification problem.

In our experiments, we compared our model against both classical and recent state-of-the-art anomaly detection methods. The classical baselines follow~\cite{Benabderrahmane21} and include AVF, FPOF, OD, OC3, VF-ARM, and VR-ARM, while the recent methods encompass {MAE-AD} (Masked AutoEncoder Anomaly Detection, as employed in MAGIC~\cite{jia2024magic}), {RT-APT}~\cite{weng2025rt}, {DeepOCCL}~\cite{wang2023deep}, {PGAE-APT}~\cite{liu2025pgae}, and {HDPT-IDS}~\cite{lee2025ml}. These more recent frameworks collectively represent modern directions in provenance-based, autoencoder-driven, and hybrid deep learning APT detection. All baselines were evaluated using the DARPA Transparent Computing datasets to ensure consistency and comparability under identical conditions. We deliberately excluded datasets that were too small, outdated, or unavailable online, as well as methods validated solely on such limited sources.

\section{ALADAEN: An Active Learning-assisted Adversarial Dual AutoEncoder for APTs Detection }

\subsection{Overview and Global Architecture}
\begin{figure}
    \centering
    \includegraphics[width=\linewidth]{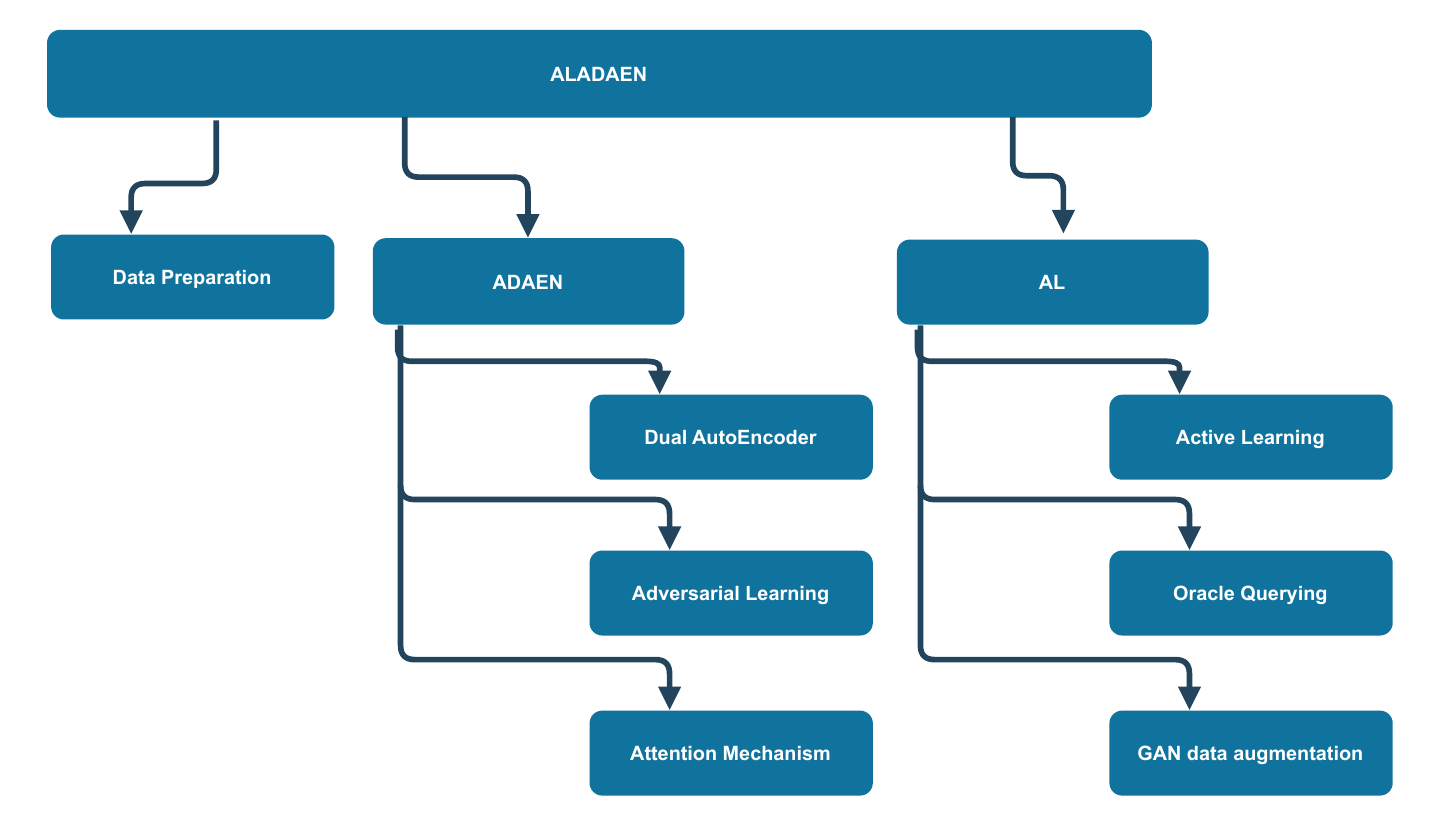}
    \caption{Overall architecture of the \textit{ALADAEN} framework: 
Active Learning-Assisted Attention Adversarial Dual AutoEncoders for anomaly detection. 
The framework is composed of three interconnected modules: 
(1) \textbf{Data Preparation}, which formats provenance events into process-level feature vectors; 
(2) \textbf{ADAEN Backbone}, a dual autoencoder with attention and adversarial training 
that learns a robust representation of benign behavior and computes reconstruction-based anomaly scores; 
and (3) \textbf{Active Learning \& GAN Augmentation}, which iteratively selects the most 
informative unlabeled samples (based on uncertainty), queries them from an oracle, 
and uses GAN-generated synthetic samples to enrich the benign pool before retraining. 
This modular design enables continuous refinement and improved anomaly ranking under scarce-label conditions.}
    \label{fig:AladaenParts}
\end{figure}

\begin{figure}
\centering
     \includegraphics[width=0.8\linewidth,height=\textheight]{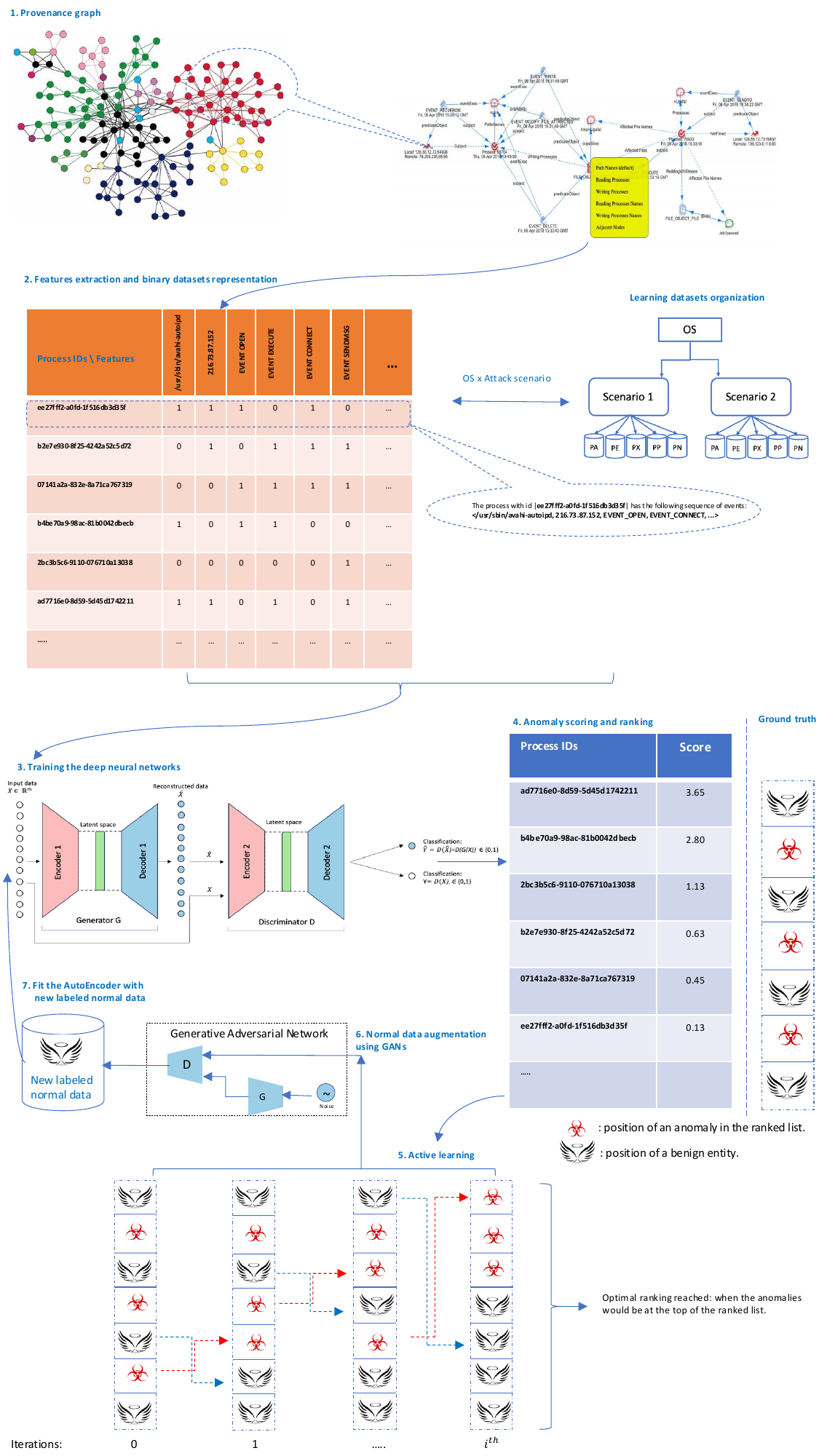}
     \caption{Step-by-step workflow of \textit{ALADAEN}. 
\textbf{Steps 1-2:} Construct process-level feature vectors from the provenance graph 
and initialize the training set with a small subset of labeled benign data (cold start). 
\textbf{Step 3:} Train the ADAEN model to reconstruct benign samples and compute 
anomaly scores based on reconstruction error. 
\textbf{Step 4:} Rank all unlabeled samples by anomaly score to prioritize potential threats. 
\textbf{Step 5:} Apply active learning to select the most uncertain samples 
(i.e., those near the decision threshold) for oracle labeling. 
\textbf{Step 6:} Augment the newly labeled benign data using a GAN to mitigate 
data scarcity and retrain the ADAEN model with the enriched dataset. 
\textbf{Step 7:} Repeat previous steps iteratively until the labeling budget is exhausted. 
This iterative cycle progressively improves anomaly ranking and detection robustness.}\label{Fig:ALADAEN}
\end{figure}

The global architecture of the \textit{ALADAEN} framework is depicted in Figure \ref{fig:AladaenParts}. It is composed of three major compartments. The first one is responsible for data formatting and preparation. The second one, called here \textit{ADAEN}, handles the training of the neural network. The third compartment, called \textit{AL}, is dedicated to active learning and GAN-based augmentation of normal labeled data.\\
The workflow of \textit{ALADAEN} consists of seven main steps, as illustrated in Figure \ref{Fig:ALADAEN} from top to bottom.  
It begins with features extraction from the provenance graph databases. The event sequences are represented row-wise in binary vectors, where each data (process) has its corresponding features (steps 1 and 2 in the figure).
After that, ADAEN AutoEncoder neural network is initially trained with a subset of normal labeled data (step 3 in the figure), aimed at learning and extracting patterns of normal behavior by minimizing the reconstruction error for these normal instances. The size of the learning subset is intentionally reduced to simulate the rarity of labeled data in real-world scenarios, showcasing the main advantage of our framework, which can iteratively improve performance despite the limited availability of labeled data. After the training, ADAEN is assessed with a test dataset (including both normal and outliers), and anomalous entities which exhibit patterns deviating from normal data will have higher reconstruction errors. Anomaly scores are then computed for each input data based on the reconstruction error, and the pairwise list [data-point, score] is sorted in a descending order. Entities with high errors are considered anomalous and are ranked accordingly. This ranking serves as a prioritized list, where processes at the top are more likely to be anomalies (step 4 in the figure). \\
The active learning component is integrated into the ALADAEN pipeline to reduce the false positive rate and improve the detection of true anomalies (step 5). The algorithm applies a threshold to identify potential anomalies based on reconstruction error. Instead of asking an oracle to blindly label all points, which is extremely costly, it selects uncertain samples above a given threshold, that are at the top of the ranked list, for further labeling. Uncertainty is measured here for data points whose reconstruction error falls near the threshold boundary, indicating that the model is uncertain about whether they are anomalous.  During each iteration of the active learning process:
(i) The system queries an oracle (e.g., a human expert) to label uncertain data points from the ranked list, specifically those near the decision boundary.
(ii) Based on the feedback, a new subset of normal labeled data is created. A synthetic dataset is then generated using a GAN, which is trained to replicate the data distribution of the normal subset. These two datasets—the real and synthetic—are merged and added back into the original training dataset to refine the model (step 6 in the figure).
(iii) Finally, the neural network is retrained on the updated dataset, progressively improving its ability to distinguish between normal and anomalous processes (step 7).

The ultimate goal is to achieve an optimal ranking quickly where all anomalies are positioned at the top of the ranked list after a few iterations. Through active learning, the model continuously improves the accuracy of the rankings by making adjustments based on new labels. The process continues until the model achieves a stable state, where further iterations do not significantly change the ranking. At this point, the model is considered optimized for anomaly detection, having learned a more accurate representation of both normal and anomalous data.

In the next subsections, we will present in details the two important parts of \textit{ALADAEN} framework, that are \textit{ADAEN} AutoEncoders, and \textit{AL} active learning combined with GANs.

\subsection{ADAEN: Attention Adversarial Dual AutoEncoder Architecture}
\subsubsection{\textbf{AutoEncoder Fundamentals}}
AutoEncoders (AE) are unsupervised neural network models that aim to learn an efficient, lower-dimensional representation (latent representations encoding) of input data by minimizing the reconstruction error between the input and output. In the context of anomaly detection, the AutoEncoder is trained to capture the patterns of normal data and can be used to detect anomalies by identifying data points with high reconstruction errors. If an input is anomalous (i.e., different from the normal data), the AutoEncoder will struggle to reconstruct it accurately, leading to a high reconstruction error. This error can then be used as an indicator of anomalies.

Let $\mathcal{X}=\{\mathbf{x}^{(i)}\}_{i=1}^{m}$ denote the dataset,
where each $\mathbf{x}^{(i)} \in \mathbb{R}^{d}$ is a $d$-dimensional feature vector
representing a process trace (network or OS activity). The AutoEncoder consists of an encoder \( f_{\theta}(\mathbf{x}^{(i)}) \) and a decoder \( g_{\theta}(\mathbf{z}^{(i)}) \), where \( \mathbf{z}^{(i)} \in \mathbb{R}^k \) is a compressed latent representation of the input. 
The \textit{Encoder} component is defined as follow:

\begin{equation}
\mathbf{z}^{(i)} = f_{\theta}\!\bigl(\mathbf{x}^{(i)}\bigr)
= \sigma\!\left(W_{e}\mathbf{x}^{(i)} + \mathbf{b}_{e}\right)
\label{eq:encoder}
\end{equation}

where \( W_e \) and \( b_e \) are the weights and biases of the encoder, \( \theta = (W_e, b_e) \), and \( \sigma \) is a non-linear activation function.
The Encoder's role is to map the original data of dimension $d$ onto a feature space of dimension $k$ where $k << d$, whereas the decoder reconstructs the original data from its encoding in the feature space.
The \textit{Decoder} reconstructs the input:
\begin{equation}
\hat{\mathbf{x}}^{(i)} = g_{\phi}\!\bigl(\mathbf{z}^{(i)}\bigr)
= \sigma\!\left(W_{d}\mathbf{z}^{(i)} + \mathbf{b}_{d}\right)
\label{eq:decoder}
\end{equation}

where \( W_d \) and \( b_d \) are the weights and biases of the decoder, and \( \phi = (W_d, b_d) \).
The AutoEncoder is trained to minimize the reconstruction error:
\begin{equation}
\mathcal{L}_{\mathrm{rec}} =
\frac{1}{|\mathcal{X}|}
\sum_{\mathbf{x}^{(i)} \in \mathcal{X}}
\left\| \mathbf{x}^{(i)} - g_{\phi}\!\bigl(f_{\theta}(\mathbf{x}^{(i)})\bigr) \right\|_2^2
\label{eq:reconstruction-loss}
\end{equation}

Through training, the two networks learn to capture the most important compressed features of the normal data points that can be used to reconstruct the data as accurately as possible. As a result, whenever presented with data points with abnormal patterns and features, the trained networks will have more difficulties in correctly reconstructing that data. Consequently, such attempts will generate larger reconstruction errors. Hence, during inference, the reconstruction error \( error(\mathbf{x}^{(i)}) = \mathcal{L}_{\mathrm{rec}} \) is computed for each input. Data points with reconstruction error above a given threshold (\( error(\mathbf{x}^{(i)}) > \tau \)) are flagged as potential anomalies.  The anomaly detection decision rule with an AutoEncoder can be written as:
\begin{equation}
\text{Anomaly}(\mathbf{x}^{(i)}) =
\begin{cases}
1 & \text{if } error(\mathbf{x}^{(i)}) \ > \tau, \\
0 & \text{otherwise}.
\end{cases}
\label{eq:anomaly-decision}
\end{equation}

The reconstruction error of a data point $\mathbf{x}^{(i)}$ after being passed through the AutoEncoder is also assigned as its anomaly score:
\begin{equation}
Ascore(\mathbf{x}^{(i)}) = error(\mathbf{x}^{(i)})
\end{equation}
\subsubsection{\textbf{ADAEN Architecture Overview}}
The Attention Adversarial Dual AutoEncoder (\textit{ADAEN}) architecture used in this paper is a variant of the baseline AutoEncoders that incorporates both adversarial training and dual autoencoding mechanisms, enhancing the detection of anomalies in complex data. It combines generative adversarial networks (GANs) with two AutoEncoders to enhance the detection of anomalies, especially in complex data distributions. The attention mechanism is applied to help the model focus on the most relevant features or timesteps of the encoded data, enhancing reconstruction and anomaly detection performance. ADAEN improves robustness and reconstruction capabilities by leveraging both a primary AutoEncoder for reconstruction and a secondary AutoEncoder to refine and balance the learning process.

\subsubsection{\textbf{Dual Adversarial Learning}}

The base of ADAEN consists of two AutoEncoders AE1 and AE2, each learning different but complementary representations of the data as represented in Figure \ref{Fig:ALADAEN} (step 3). By definition:

\begin{itemize}
    \item {Primary AutoEncoder (AE1):} 
    Focuses on reconstructing the input data (normal instances) by minimizing the reconstruction error. 
    For each input sample $\mathbf{x}^{(i)}$:
    \begin{align*}
        \text{Encoder 1:} \quad & 
        \mathbf{z}_{1}^{(i)} = f_{\theta_1}\!\bigl(\mathbf{x}^{(i)}\bigr) \\
        \text{Decoder 1:} \quad & 
        \hat{\mathbf{x}}_{1}^{(i)} = g_{\phi_1}\!\bigl(\mathbf{z}_{1}^{(i)}\bigr)
    \end{align*}

    \item {Secondary AutoEncoder (AE2):}
    Works alongside AE1 to refine the reconstruction and capture complementary representations. 
    For the same input sample $\mathbf{x}^{(i)}$:
    \begin{align*}
        \text{Encoder 2:} \quad & 
        \mathbf{z}_{2}^{(i)} = f_{\theta_2}\!\bigl(\mathbf{x}^{(i)}\bigr) \\
        \text{Decoder 2:} \quad & 
        \hat{\mathbf{x}}_{2}^{(i)} = g_{\phi_2}\!\bigl(\mathbf{z}_{2}^{(i)}\bigr)
    \end{align*}
\end{itemize}

These two AutoEncoders help capture different but complementary representations of the data, which improves the robustness of the system.
\noindent
The reconstruction losses for the two AutoEncoders are defined as:
\begin{equation}
\mathcal{L}_{\mathrm{rec}_1} =
\frac{1}{|\mathcal{X}|}
\sum_{\mathbf{x}^{(i)} \in \mathcal{X}}
\left\| \mathbf{x}^{(i)} - 
g_{\phi_1}\!\bigl(f_{\theta_1}(\mathbf{x}^{(i)})\bigr)
\right\|_2^2
\label{eq:reconstruction-loss-ae1}
\end{equation}
\begin{equation}
\mathcal{L}_{\mathrm{rec}_2} =
\frac{1}{|\mathcal{X}|}
\sum_{\mathbf{x}^{(i)} \in \mathcal{X}}
\left\| \mathbf{x}^{(i)} - 
g_{\phi_2}\!\bigl(f_{\theta_2}(\mathbf{x}^{(i)})\bigr)
\right\|_2^2
\label{eq:reconstruction-loss-ae2}
\end{equation}

The total reconstruction error for the ADAEN is a weighted combination of these two errors:
\begin{equation}
\mathcal{L}_{\mathrm{rec}} =
\alpha \mathcal{L}_{\mathrm{rec}_1} +
(1-\alpha)\mathcal{L}_{\mathrm{rec}_2},
\quad \alpha \in [0,1]
\label{eq:combined-reconstruction-loss}
\end{equation}

where \( \alpha \in [0,1] \) is a hyper parameter that balances the contributions of the two AutoEncoders.
\subsubsection{\textbf{Adversarial Component}}

ADAEN integrates adversarial training, inspired by GANs, to enhance the model's ability to capture the data distribution. The discriminator \( D_{\psi} \) is defined through the second AutoEncoder, which is trained to distinguish between real data samples and the reconstructed samples produced by the first AutoEncoder that plays the role of the generator (Figure \ref{Fig:ALADAEN}, step 3).

The discriminator \( D_{\psi}\) assigns a probability that a sample \( \mathbf{x}^{(i)} \) is real (i.e., from the true data distribution), and helps improve the first AutoEncoder to generate more realistic reconstructions that closely resemble the original data.
\noindent
The adversarial loss for the discriminator is given by:
\begin{align}
\mathcal{L}_{D} &=
- \mathbb{E}_{\mathbf{x}^{(i)} \sim p_{\text{data}}}
    \bigl[\log D_{\psi}(\mathbf{x}^{(i)})\bigr] \nonumber \\
&\quad
- \mathbb{E}_{\hat{\mathbf{x}}_1^{(i)}, \hat{\mathbf{x}}_2^{(i)} \sim p_{\text{reconstruction}}}
    \!\bigl[
        \log\!\bigl(1 - D_{\psi}(\hat{\mathbf{x}}_1^{(i)})\bigr) +
        \log\!\bigl(1 - D_{\psi}(\hat{\mathbf{x}}_2^{(i)})\bigr)
    \bigr]
\label{eq:discriminator-loss}
\end{align}

The generator (AE1) is trained to fool the discriminator by making its reconstructions indistinguishable from real samples. 
Its adversarial loss is:
\begin{equation}
\mathcal{L}_{\mathrm{adv}} =
- \mathbb{E}_{\hat{\mathbf{x}}_1^{(i)}, \hat{\mathbf{x}}_2^{(i)} \sim p_{\text{reconstruction}}}
    \!\bigl[
        \log D_{\psi}(\hat{\mathbf{x}}_1^{(i)}) +
        \log D_{\psi}(\hat{\mathbf{x}}_2^{(i)})
    \bigr]
\label{eq:adversarial-loss}
\end{equation}

Finally, the overall objective for ADAEN combines reconstruction and adversarial terms:
\begin{equation}
\mathcal{L}_{\mathrm{ADAEN}} =
\mathcal{L}_{\mathrm{rec}} +
\lambda \mathcal{L}_{\mathrm{adv}},
\quad \lambda > 0
\label{eq:adaen-loss}
\end{equation}

where $\lambda > 0$ is a weighting hyperparameter that balances 
the contribution of the reconstruction and adversarial losses.
In our experiments, we set $\lambda = 0.5$ as a simple and effective choice.\\
Training ADAEN involves the following steps:

\begin{enumerate}
    \item {Train Discriminator}: Update the discriminator \( D_{\psi} \) to minimize the adversarial loss \( \mathcal{L}_{D} \).
  
    \item {Train AutoEncoders}: Update the parameters of both AutoEncoders \( \theta_1, \phi_1, \theta_2, \phi_2 \) to minimize the combined loss \( \mathcal{L}_{\text{ADAEN}} \).
\end{enumerate}

The training alternates between updating the discriminator and the AutoEncoders, that is similar to the GAN training process.

When applied to anomaly detection, ADAEN uses both AutoEncoders to compute the reconstructions, which are then compared to detect anomalies.

As in the case of a baseline AutoEncoder, for a given input sample 
$\mathbf{x}^{(i)}$, the final anomaly score is computed as the 
combined reconstruction loss:
\begin{equation}
Anomaly(\mathbf{x}^{(i)}) = 
\mathcal{L}_{\mathrm{rec}}\!\bigl(\mathbf{x}^{(i)}\bigr)
= \alpha \mathcal{L}_{\mathrm{rec}_1}\!\bigl(\mathbf{x}^{(i)}\bigr) +
(1-\alpha)\mathcal{L}_{\mathrm{rec}_2}\!\bigl(\mathbf{x}^{(i)}\bigr)
\label{eq:anomaly-score-dual}
\end{equation}

If $Anomaly(\mathbf{x}^{(i)}) > \tau$, the input is classified as an anomaly:
\begin{equation}
\text{Anomaly}(\mathbf{x}^{(i)}) =
\begin{cases}
1 & \text{if } Anomaly(\mathbf{x}^{(i)}) > \tau, \\
0 & \text{otherwise}.
\end{cases}
\label{eq:anomaly-decision-dual}
\end{equation}

\subsubsection{\textbf{Attention Mechanism}}

To enhance the generator's reconstruction (AE1), an extra \textit{attention mechanism} layer is added to the encoded features through Encoder 1 component. Its role is to help the model focus on key parts of the input data during the generation process, ensuring that important features are captured more effectively while generating synthetic data. This would help the generator create higher-quality reconstructions of normal data and better identify patterns, which is crucial when generating new synthetic data.

Given Encoder 1’s output \( z \in \mathbb{R}^{T \times k} \) (where \( T \) is the number of timesteps and \(k \) is the feature dimension in the latent space), the attention mechanism computes an attention score \( \xi_i \) for each timestep \( i \), helping the model focus on important timesteps.

The \textit{attention score} \( \xi_i \) is computed as:
\[
\xi_i = \frac{\exp(e_i)}{\sum_{j=1}^T \exp(e_j)}
\]
where \( e_i \) is the relevance score for timestep \( i \), typically computed as a dot product of the hidden state \( h_i \) and a learned context vector \( v \):
\[
e_i = v^\top h_i
\]

The context vector \( c \) is the weighted sum of the Encoder 1 outputs:
\[
c = \sum_{i=1}^{T} \xi_i h_i
\]
This context vector \( c \) is used by the Decoder 1 to reconstruct the input.

\subsubsection{\textbf{Rationale for Dual AutoEncoder Design}}
\label{subsec:dualAE-discussion}

The choice of a dual AutoEncoder (AE) architecture with an adversarial component and an attention mechanism in ADAEN is motivated by the need to model highly complex and heterogeneous behaviors in cybersecurity provenance data. This design provides the following conceptual advantages:  

\begin{itemize}
    \item {Enhanced data distribution modeling:} Using two AutoEncoders allows the system to learn complementary latent representations, offering a richer description of normal behavior and improving sensitivity to subtle deviations.
    \item {Error refinement through adversarial consistency:} AE2 acts as a reconstruction-refiner for AE1, learning to produce outputs that are closer to the normal manifold. Anomalies that AE1 reconstructs too well still incur disagreement with AE2, which amplifies their anomaly scores.
    \item {Robustness through redundancy:} The presence of two reconstruction paths introduces diversity in learned representations, reducing the chance that both models will underfit or overfit the same features simultaneously.
    \item {Generalization and stability:} The adversarial objective forces AE1 to improve reconstructions iteratively so that they cannot be distinguished by AE2 as abnormal, encouraging better generalization and reducing overfitting.
    \item {Attention-driven focus on key events:} The attention mechanism highlights the most informative features of the input, improving reconstruction quality where it matters most for anomaly detection.
\end{itemize}

Together, these design elements allow ADAEN to better separate normal and anomalous behaviors, particularly under the severe class imbalance characteristic of APT detection. The resulting anomaly scores are more robust, stable, and aligned with operational needs, providing a principled foundation for the active learning loop and subsequent ranking-based triage.

To provide reproducibility and enable direct comparison with future work, we next summarize the detailed architecture of ADAEN and the GAN module employed.
\subsubsection{Architectural Details}
Each AutoEncoder (AE1 and AE2) uses a fully-connected feedforward architecture
whose layer widths are defined relative to the input dimension $d$ 
(number of features per process). 
The encoder consists of three layers with widths 
$[d \rightarrow d/2 \rightarrow d/4 \rightarrow k]$, 
where $k=32$ is the latent dimension. 
The decoder is symmetric with widths 
$[k \rightarrow d/4 \rightarrow d/2 \rightarrow d]$. 
All hidden layers use LeakyReLU activation ($\alpha=0.2$), 
followed by batch normalization. 
A dropout layer ($p=0.2$) is applied after the first hidden layer 
to prevent overfitting.

For the adversarial component, the discriminator $D_{\psi}$ is a 
three-layer MLP with widths $[d \rightarrow d/2 \rightarrow d/4 \rightarrow 1]$, 
where the last layer outputs a probability score via a sigmoid activation.
The generator is AE1, trained to minimize the non-saturating GAN loss.
Both AE1/AE2 and $D_{\psi}$ are optimized using Adam 
($\beta_1=0.5, \beta_2=0.999$, learning rate $10^{-4}$).


\subsection{Active Learning}
\subsubsection{\textbf{Background}}
Active learning is a paradigm where the model selectively queries the user (or an oracle) to label the most informative examples, aiming to improve the model with fewer labeled examples.
This approach is especially useful when labeling data is costly or time-consuming which is the case with APTs where the learning databases are rare. The model focuses on uncertain or difficult examples that can help refine its accuracy.   
Intuitively, the motivation behind our utilization of active learning for APTs detection comes from rewards and penalties in cybersecurity. In such a context, positive
rewards are typically given when there is a correct detection of an attack (e.g., DDoS, malware, phishing) by flagging the traffic as anomalous or malicious.
The model could also consider a positive reward during a successful mitigation of an attack, i.e. if the relevant action is taken to successfully block an attack or reduce its impact (e.g., blocking malicious IPs, isolating infected systems). Inversely, a penalty would happen when the model fails to protect the system or causes unnecessary disruption. In fact, if the model incorrectly flags normal behavior as malicious, it will incur a penalty. For instance, blocking legitimate user access or flagging benign network traffic as malicious would cause the system to be disrupted, leading to a penalty.
 
\subsubsection{\textbf{Oracle Querying}}

After the initial training of \textit{ADAEN} neural network on a dataset primarily consisting of normal data, the goal is for the model to learn how to reconstruct normal patterns with minimal error.
The framework passes the test data through the AutoEncoder and calculates the reconstruction error for each example. Then it uses a threshold on reconstruction error to flag potential anomalies (points with a high reconstruction error).\\
Instead of manually labeling a large number of data points, the framework uses active learning strategy to query the most informative samples. These could be: points near the threshold of reconstruction error (uncertain if they are anomalous or not); or points with the highest reconstruction error, likely to be anomalous. During the active learning loop, the most uncertain data points are identifier for labeling and the uncertainty of a sample \( x \) is quantified by its proximity to the decision boundary. Mathematically, this is often represented using the model's confidence score: $U(x) = 1 - p(\hat{y}|x)$ where \( p(\hat{y}|x) \) is the predicted probability for the most likely class label \( \hat{y} \). Data points with the highest uncertainty (lowest confidence) are selected for labeling. Then the framework queries an oracle (e.g., a human expert) to label these points as normal or anomalous.\\
After obtaining the labels for these ambiguous points from an oracle, the framework uses them to train a Generative Neural Network (GAN), and a synthetic dataset is produced by replicating the data distribution of the newly labeled ambiguous points. Both datasets are then merged and used to refine the AutoEncoder’s threshold for anomaly detection and retrain it with the newly labeled data.

\subsubsection{\textbf{Ambiguous Points Augmentation with GAN}}
We employ a Generative Adversarial Network (GAN) to augment the ambiguous (uncertain) data points during the active learning process. Specifically, once the most uncertain or ambiguous data points are identified and labeled by the oracle, a GAN is used to generate synthetic data that closely mimics the distribution of the normal labeled data. The uncertain points labeled as normal by the oracle are then combined with the synthetic data generated by the GAN, and this augmented dataset is used to retrain the AutoEncoders. This augmentation helps to enrich the dataset, especially in scenarios where labeled data is scarce. 
This should also improve the reconstruction capabilities of the AutoEncoders for normal data, leading to better detection of anomalies.
By iteratively incorporating synthetic samples generated by the GAN into the training process, we enhance the model's capacity to learn from limited data and improve its performance in distinguishing between normal and anomalous instances. This strategy is particularly effective in addressing the imbalance typically seen in anomaly detection tasks.\\
The used GAN consists of two neural networks, a generator \( G \) and a discriminator \( D \), that are trained together in a competitive process. Both are defined as follows:

The Generator \( G \): Given a random noise vector \( z \sim p_z(z) \), the generator tries to produce synthetic data \( G(z) \) that resembles the distribution of the uncertain points labeled as normal  \( p_{\text{data}}(x) \). Thus:
\[
G: z \mapsto G(z) \quad \text{where} \quad z \sim \mathcal{N}(0, I)
\]

The Discriminator \( D \): is a classifier that attempts to distinguish between real data samples \( x \sim p_{\text{data}}(x) \) and synthetic data \( G(z) \). The output of the discriminator is a probability \( D(x) \) that indicates how likely a given sample is real:

\[
D: x \mapsto [0, 1] \quad \text{where} \quad D(x) \text{ is the probability that } x \text{ is real.}
\]

The training objective is a minimax optimization problem:

\[
\min_G \max_D \mathbb{E}_{x \sim p_{\text{data}}(x)}[\log D(x)] + \mathbb{E}_{z \sim p_z(z)}[\log(1 - D(G(z)))]
\] 

\subsubsection{\textbf{Iterative Loop}}
The whole process can be repeated iteratively: by reapplying the AutoEncoder to the augmented dataset, and using active learning to query additional informative samples. Over time, the model becomes better at distinguishing between normal and anomalous points, and the detection rate improves with fewer false positives and negatives.

There are several key benefits behind the utilization of active learning within our anomaly detection framework. For instance, it ensures an efficient labeling by minimizing the number of data that need to be labeled (hence the cost), focusing on the most ambiguous or informative points. This is particularly helpful when anomalies are rare and hard to label. Through the iterative learning, the model gets progressively better as more labeled data becomes available, without the need for massive amounts of labeled training data at the start. 

The sampling strategies in the Active Learning process could be diverse. For example, with uncertainty sampling, the model selects data points that are close to the decision boundary for labeling (i.e., points with reconstruction errors near the anomaly threshold). These points are the most uncertain, and labeling them helps refine the boundary between normal and anomalous data.
With diversity sampling, the model selects diverse points that cover different regions of the data space to ensure that the model learns about various types of potential anomalies. With outlier sampling, the model selects points with the highest reconstruction error, as they are most likely to be anomalous. In our case, the outlier and uncertainty sampling have been chosen since they are more appropriate to APTs detection.

\subsection{Anomaly Ranking and Evaluation metrics}

In Advanced Persistent Threat (APT) detection, where malicious processes typically comprise less than 0.1\% of the dataset, relying on accuracy as a performance metric is misleading. A model could trivially attain near-perfect accuracy by labeling all processes as benign, thereby failing to detect actual threats. Consequently, we shift our focus from overall accuracy to the model’s ability to identify and prioritize APT-related anomalies.

This reframes the task as a ranking problem: instead of binary labels ("normal" vs. "anomalous"), the model assigns a score to each process that reflects its likelihood of being anomalous. The resulting ranked list allows security analysts to review the most suspicious processes first—a setting that better aligns with operational cybersecurity workflows. Ranking is also well-suited to active learning, where the model iteratively selects top-scored instances (i.e., most uncertain or anomalous) for expert labeling, thus accelerating the learning process with minimal supervision.

To evaluate the quality of this ranking, we use the Normalized Discounted Cumulative Gain (nDCG) \cite{jarvelin_2002}, a widely adopted metric in information retrieval. nDCG assesses how effectively a model places the truly anomalous instances near the top of the ranked list \cite{BerradaCBMMTW20, Benabderrahmane21}. The score ranges from 0 to 1, with 1 indicating that all anomalies are perfectly ranked at the top, regardless of their internal order.

First, the Discounted Cumulative Gain (DCG) is computed as:

\begin{equation}
DCG = \sum_{i=1}^{N} \frac{rel_i}{\log_2(i + 1)},
\end{equation}

where $rel_i$ denotes the binary relevance (1 for anomaly, 0 otherwise) of the $i^\text{th}$ item in the ranked list of $N$ processes. The DCG is then normalized by the ideal DCG (iDCG), which corresponds to a perfect ranking:

\begin{equation}
nDCG = \frac{DCG}{iDCG}.
\end{equation}

In our setting, each process receives an anomaly score, and the nDCG evaluates how well true anomalies are ranked above benign processes. Maximum nDCG is achieved when all anomalous instances occupy the top $k$ positions (where $k$ is the number of anomalies), irrespective of their mutual order.

This ranking-based evaluation provides a robust and interpretable measure of detection performance, especially in highly imbalanced scenarios typical of APT campaigns, where the objective is to surface rare but critical threats.
\subsection{ALADAEN Pseudo-code}

Algorithm 1 gives the pseudo-code of ALADAEN framework. It iteratively trains the ADAEN AutoEncoder model, queries an oracle for labels, and refines the model using new labeled data. The algorithm uses a ranking system based on reconstruction errors and uncertainty scores to identify anomalies and enhance detection performance.
\begin{algorithm}
\caption{ALADAEN(): Active Learning-Assisted Attention Adversarial Dual AutoEncoders}
\scriptsize
\begin{algorithmic}[1]
\STATE \textbf{Input:} 
\begin{itemize}
    \item Labeled dataset $D_L$
    \item Unlabeled dataset $D_U$
    \item Query budget $Q$ (number of samples to query per iteration)
    \item Threshold $\tau$ for anomaly detection
    \item Oracle for labeling queried samples
    \item Ground truth anomaly labels $\mathcal{Y}_{\text{ground truth}}$ for evaluation
      \item Number of iterations $N_{\text{iterations}}$
\end{itemize}


\STATE \textbf{Step 1: AutoEncoder-Based Anomaly Detection}
\STATE Train ADAEN on the labeled dataset $D_L$ to capture normal data patterns.
\STATE Compute reconstruction error for each sample in the unlabeled dataset $D_U$.
\FOR{each sample $\mathbf{x}^{(i)} \in D_U$}
    \STATE Compute the reconstruction error: $\text{error}(\mathbf{x}^{(i)}) = ||\mathbf{x}^{(i)} - \text{ADAEN}(\mathbf{x}^{(i)})||_{2}^2 $
    \STATE Compute the anomaly score:$
    S(\mathbf{x}^{(i)}) = \mathcal{L}_{\text{rec}}(\mathbf{x}^{(i)})$
    
\STATE If $S(\mathbf{x}^{(i)}) > \tau$, classify $\mathbf{x}^{(i)}$ as an anomaly:
\[
\text{Anomaly}(\mathbf{x}^{(i)}) = 
\begin{cases} 
1 & \text{if } S(\mathbf{x}^{(i)}) > \tau \\
0 & \text{otherwise}
\end{cases}
\]

\ENDFOR

\STATE \textbf{Step 2: Active Learning Loop}
\REPEAT
 \STATE Sort the anomaly scores $S(\mathbf{x}^{(i)})$ in descending order: $
    S_{\text{sorted}} = \text{sort}(S(\mathbf{x}^{(i)}))$
    \STATE Compute uncertainty scores for each sample $x \in D_U$ based on proximity to threshold $\tau$:
    \[
    U(x) = | S_{\text{sorted}}- \tau |
    \]

    \STATE \textbf{Step 3: Compute nDCG Score}
    \STATE Match the sorted scores with ground truth labels $\mathcal{Y}_{\text{ground truth}}$ and calculate nDCG.
    \[
    nDCG_{\text{score}}=nDCG(S_{\text{sorted}},\mathcal{Y}_{\text{ground truth}})
    \]

        \STATE Print the nDCG for the current iteration: $Print(nDCG_{\text{score}})$

      \STATE \textbf{Step 4: Querying}
    \STATE Select top $Q$ samples with the highest uncertainty scores:
    \[
    \{ x_1, x_2, \dots, x_{Q} \} = \arg \max_{\mathbf{x}^{(i)} \in D_U} U(\mathbf{x}^{(i)})
    \]
    
    \STATE Query the oracle for the true labels of the selected samples:
    \[
    \mathcal{Y}_{\text{uncertain}} = \text{Oracle}(\{ x_1, x_2, \dots, x_{Q} \})
    \]
    \STATE Use a GAN for data augmentation: $\hat{\mathcal{Y}}_{\text{uncertain}}=GAN(
    \mathcal{Y}_{\text{uncertain}}).$
    
    \STATE Add labeled data to $D_L \leftarrow D_L \cup \mathcal{Y}_{\text{uncertain}} \cup \hat{\mathcal{Y}}_{\text{uncertain}}$ 

    
    \STATE Remove queried samples from the unlabeled set $D_U$:
    \[
    D_U \leftarrow D_U \setminus \mathcal{Y}_{\text{uncertain}}
    \]
    
    \STATE \textbf{Step 3: Retrain AutoEncoder with Newly Labeled Data}
    \STATE Retrain AutoEncoder on the combined labeled dataset $D_L$:
    \[
    \mathcal{L}_{\text{reconstruction}} = \frac{1}{|D_L|} \sum_{\mathbf{x}^{(i)} \in D_L} || \mathbf{x}^{(i)} - \hat{\mathbf{x}^{(i)}} ||^2
    \]
    
    \FOR{each sample $\mathbf{x}^{(i)} \in D_U$}
        \STATE Recompute reconstruction error and uncertainty score for $\mathbf{x}^{(i)}$.
    \ENDFOR

\UNTIL{query budget $Q$ is exhausted or $N_{\text{iterations}}$ stopping criterion is met}


\STATE \textbf{Output:} 
\begin{itemize}
    \item Trained AutoEncoder model
    \item Updated labeled dataset $D_L$
    \item Final anomaly detection results
    \item Final nDCG score.
\end{itemize}
\label{algALADAEN}
\end{algorithmic}
\end{algorithm}
%
%
%
\subsubsection{\textbf{Step 1: AutoEncoder Training}}
\begin{itemize}
    \item \textbf{Objective}: Train the Attention Adversarial Dual AutoEncoder (ADAEN) model on a labeled sub-dataset to capture the normal data patterns and detect anomalies.
    \item The algorithm begins by training the ADAEN model on the labeled dataset $D_L$, which consists of known normal data. The model is trained to minimize the reconstruction loss between the original data and the reconstructed data generated by the AutoEncoder.
    \item \textbf{Reconstruction Error Calculation}: After training, the algorithm computes the reconstruction error for each sample in the unlabeled dataset $D_U$. 
    The reconstruction error for a sample $\mathbf{x}^{(i)}$ is computed as:
\[
\text{error}\!\bigl(\mathbf{x}^{(i)}\bigr) =
\left\|
\mathbf{x}^{(i)} -
\text{ADAEN}\!\bigl(\mathbf{x}^{(i)}\bigr)
\right\|_2^2,
\]
where $\text{ADAEN}(\cdot)$ denotes the combined reconstruction produced by the dual autoencoder, i.e.,
\[
\text{ADAEN}\!\bigl(\mathbf{x}^{(i)}\bigr) =
\alpha g_{\phi_1}\!\bigl(f_{\theta_1}(\mathbf{x}^{(i)})\bigr) +
(1-\alpha)g_{\phi_2}\!\bigl(f_{\theta_2}(\mathbf{x}^{(i)})\bigr).
\]
    \item \textbf{Anomaly Scoring}: The reconstruction error is then used to compute the anomaly score for each sample:
\[
S(\mathbf{x}^{(i)}) =
\mathcal{L}_{\mathrm{rec}}\!\bigl(\mathbf{x}^{(i)}\bigr)
= \alpha \mathcal{L}_{\mathrm{rec}_1}\!\bigl(\mathbf{x}^{(i)}\bigr) +
(1-\alpha)\mathcal{L}_{\mathrm{rec}_2}\!\bigl(\mathbf{x}^{(i)}\bigr).
\]

    Samples with high reconstruction errors are more likely to be anomalies.
    \item \textbf{Anomaly Classification}: Each sample is classified as an anomaly if its anomaly score exceeds a predefined threshold $ \tau $:
    \[
    \text{Anomaly}(\mathbf{x}^{(i)}) = 
    \begin{cases} 
    1 & \text{if } S(\mathbf{x}^{(i)}) > \tau \\
    0 & \text{otherwise}
    \end{cases}
    \]
\end{itemize}
The choice of the threshold values is explained in the experiments section.
\subsubsection{\textbf{Step 2: Active Learning Loop}}
\begin{itemize}
    \item \textbf{Objective}: Iteratively improve the model's performance by querying the oracle for labels of uncertain samples and retraining the model with the newly labeled data.
    \item \textbf{Anomaly Score Sorting}: The anomaly scores are then sorted in descending order to prioritize the 
most anomalous samples for querying:
\[
\mathcal{S}_{\mathrm{sorted}} =
\operatorname{sort}\!\bigl(\{S(\mathbf{x}^{(i)}) \mid 
\mathbf{x}^{(i)} \in D_{U}\}\bigr),
\quad \text{descending}.
\]
\item \textbf{Uncertainty Calculation:}
For each sample $\mathbf{x}^{(i)} \in D_{U}$, 
the uncertainty score is computed as the distance between its anomaly score 
and the decision threshold $\tau$:
\[
U(\mathbf{x}^{(i)}) =
\bigl| S(\mathbf{x}^{(i)}) - \tau \bigr|.
\]
Samples with $U(\mathbf{x}^{(i)})$ close to zero are considered the most uncertain 
and are prioritized for oracle querying.

In ALADAEN, the threshold $\tau$ is initialized based on the $q$-th percentile 
of the reconstruction errors in the current pool:
\[
\tau = 
\operatorname{Percentile}\!\left(
\{ s(\mathbf{x}^{(i)}) \mid \mathbf{x}^{(i)} \in 
D_{U} \}, q
\right),
\]
where $q \in [0,100]$ is a user-defined percentile 
(e.g., $q=80$ for the 80th percentile). 
Samples with $s(\mathbf{x}^{(i)}) > \tau$ are flagged as potential anomalies.

%
%
    
\end{itemize}

\subsubsection{\textbf{Step 3: Compute nDCG Score}}
\begin{itemize}
    \item \textbf{Objective}: Evaluate the quality of the ranking of anomaly scores using the normalized discounted cumulative gain (nDCG) metric.
    \item The nDCG score is computed by matching the sorted anomaly scores $ S_{\text{sorted}} $ with the ground truth labels $ \mathcal{Y}_{\text{ground truth}} $:
    \[
    {nDCG}_{\text{score}}=nDCG(S_{\text{sorted}},\mathcal{Y}_{\text{ground truth}})
    \]
    The nDCG score reflects how well the model ranks the true anomalies at the top of the list, with a score close to 1 indicating a perfect ranking.
\end{itemize}

\subsubsection{\textbf{Step 4: Querying the Oracle}}
\begin{itemize}
    \item \textbf{Objective}: Select the most uncertain samples and query the oracle (e.g., a human expert) to obtain their true labels.
    \item The algorithm selects the top $Q$ samples with the highest uncertainty scores:
    \[
    \{ x_1, x_2, \dots, x_Q \} = \arg \max_{\mathbf{x}^{(i)} \in D_U} U(\mathbf{x}^{(i)})
    \]
    \item \textbf{Oracle Interaction}: The oracle provides the true labels for the selected samples:
    \[
    \mathcal{Y}_{\text{uncertain}} = \text{Oracle}(\{ x_1, x_2, \dots, x_Q \})
    \]
    \item \textbf{Data augmentation with GAN:} 
    A generative neural network is trained with the newly labeled samples $\mathcal{Y}_{\text{uncertain}}$ to produce a similar synthetic dataset:
    \[
    \hat{\mathcal{Y}}_{\text{uncertain}}=GAN(
    \mathcal{Y}_{\text{uncertain}}). \]
    Both $\mathcal{Y}_{\text{uncertain}}$  and $\hat{\mathcal{Y}}_{\text{uncertain}}$ are combined, and added to the labeled dataset $ D_L $ for further training:
    \[
    D_L \leftarrow D_L \cup \mathcal{Y}_{\text{uncertain}} \cup \hat{\mathcal{Y}}_{\text{uncertain}}
    \]
\end{itemize}

\subsubsection{\textbf{Step 5: Retraining the AutoEncoder}}
\begin{itemize}
    \item \textbf{Objective}: Retrain the ADAEN model on the updated labeled dataset that includes both the initial labels and the new labels from the oracle.
    \item After updating $ D_L $, the AutoEncoder is retrained to minimize the reconstruction loss:
    \[
    \mathcal{L}_{\text{reconstruction}} = \frac{1}{|D_L|} \sum_{\mathbf{x}^{(i)} \in D_L} || \mathbf{x}^{(i)} - \hat{\mathbf{x}^{(i)}} ||^2
    \]
    \item The retraining helps the model refine its understanding of both normal and anomalous data patterns, improving its ability to detect anomalies in future iterations.
\end{itemize}

\subsubsection{\textbf{Termination}}
The algorithm repeats the active learning loop until the query budget $ Q $ is exhausted or the predefined stopping criterion ($N_{\text{iterations}}$) is met. Once the loop terminates, the final trained AutoEncoder and the results (e.g., final nDCG scores) are output.

\subsubsection{\textbf{Output}}
The final output includes:
\begin{itemize}
    \item The trained AutoEncoder model
    \item The updated labeled dataset $D_L$
    \item The final anomaly detection results
    \item The final nDCG scores, which indicate the quality of the anomaly ranking.
\end{itemize}

\subsection{Contribution Summary and Distinction from Prior Work}
\label{sec:novelty}

The proposed ALADAEN framework introduces several distinctive design choices that set it apart from existing anomaly detection methods that combine AutoEncoders with adversarial training or active learning. 

First, unlike adversarially trained AutoEncoder approaches such as ALAD~\cite{zenati2018efficient}, GANomaly~\cite{akcay2019ganomaly}, and ALOCC~\cite{sabokrou2018adversarially}, which couple a single AutoEncoder with a separate discriminator network, ALADAEN employs a \emph{dual-AutoEncoder adversarial design}. In this setting, AE1 acts as a generator while AE2 functions as a reconstruction-refiner/discriminator, providing complementary latent representations and disagreement-aware reconstructions that sharpen anomaly signals under severe class imbalance.

Second, ALADAEN integrates an \emph{attention layer} in the encoder path, allowing the model to focus on salient event/time steps within long and sparse provenance-derived behavioral sentences. To the best of our knowledge, this is the first time attention has been explicitly applied to provenance event sequences in an adversarial dual-AutoEncoder context.

Third, whereas prior active learning methods for anomaly detection typically rely on classifier-margin or uncertainty sampling and evaluate performance using AUROC or accuracy, ALADAEN adopts a \emph{ranking-centric active learning loop}. We select near-threshold reconstruction-uncertain processes and directly optimize the nDCG metric, aligning the learning process with the operational requirements of Security Operations Center (SOC) workflows where analysts must prioritize top-ranked threats under strict time constraints.

Fourth, ALADAEN performs \emph{GAN-based augmentation} exclusively on the oracle-confirmed normal pool obtained during active learning. This densifies the normal manifold without introducing synthetic anomalies that might shift the decision boundary, making the approach robust under extreme data scarcity.

Finally, we demonstrate the operational scope of ALADAEN through a comprehensive evaluation on \emph{40 heterogeneous datasets} (4 OS $\times$ 2 scenarios $\times$ 5 views) from the DARPA Transparent Computing (TC) program. Prior work such as USAD~\cite{audibert2020usad} primarily targets multivariate time series and does not integrate attention, GAN-based augmentation, or ranking-driven active learning. 

\section{Experimental settings, results and analysis}
In the following sections, we present the datasets and the benchmarking protocol for the evaluation of ALADAEN against the previously mentioned state-of-the-art anomaly detection methods, including classical baselines (AVF~\cite{koufakou_2007, BerradaCBMMTW20}, OC3~\cite{smets2011}, OD~\cite{narita_2008}, FPOF~\cite{he_2005}, VR-ARM~\cite{Benabderrahmane21}, VF-ARM~\cite{Benabderrahmane21}) and recent deep learning-based approaches such as {MAE-AD} (Masked AutoEncoder Anomaly Detection, as employed in MAGIC~\cite{jia2024magic}), {RT-APT}~\cite{weng2025rt}, {DeepOCCL}~\cite{wang2023deep}, {PGAE-APT}~\cite{liu2025pgae}, and {HDPT-IDS}~\cite{lee2025ml}.
%

For an initial comparison, we utilize the full training datasets without integrating active learning into ALADAEN. This evaluation allows us to benchmark ALADAEN's core architecture against other established methods and assess its effectiveness in detecting anomalies within complex data environments. Following this, we will evaluate the impact of the active learning mechanism on ALADAEN's performance. By iteratively querying an oracle, we aim to demonstrate the contribution of active learning in quickly improving detection rates and reducing reliance on large labeled datasets.

While the compared methods are diverse in their approaches to anomaly detection, our primary objective is not to just compare ALADAEN with these techniques. Rather, our main focus is to demonstrate how ALADAEN, by incorporating active learning, can significantly enhance anomaly detection performance using only a small labeled dataset.
The key insight of our work is that by iteratively querying an oracle for the labels of the most uncertain samples in the dataset, we can improve the model's understanding of both normal and anomalous data. This active learning loop allows us to maximize the utility of a small amount of labeled data, gradually refining the AutoEncoder model's ability to detect anomalies with each iteration.
Thus, the comparison with other methods serves primarily as a baseline, providing the context for the improvements that can be achieved by integrating active learning into the anomaly detection process. Our experiments aim to show that even with a limited initial labeled dataset—which is the case for APT—ALADAEN can iteratively improve its performance and achieve competitive or superior results in comparison to other state-of-the-art methods. 

In the following subsections, we present the experimental setup, describe the datasets used, and discuss the results in detail. 

\subsection{Datasets}\label{datasets}
The datasets used in this study\footnote{\url{https://gitlab.com/adaptdata}} originate from DARPA’s \verb|Transparent Computing TC| program~\footnote{\url{https://www.darpa.mil/program/transparent-computing}}, which was designed to capture comprehensive system provenance across diverse operating systems and software layers. These datasets encompass system-level activities, background operations, and execution traces recorded during simulated APT-style attacks. By preserving detailed provenance information, the data enables fine-grained tracking of interactions and dependencies between system components. This holistic view facilitates the identification of behaviors that may appear benign in isolation but collectively reveal signs of abnormal or malicious activity.\\
More specifically, we utilize here a curated subset of DARPA’s provenance data that has been preprocessed by the \verb|ADAPT| (Automatic Detection of Advanced Persistent Threats) project’s ingestion pipeline~\cite{berrada2019, BerradaCBMMTW20, Benabderrahmane21}. The datasets originate from four distinct operating systems: Android (referred to as Clearscope), Linux (Trace), BSD (Cadets), and Windows (Fivedirections or 5dir). For each OS, the data includes traces from two different attack scenarios: Scenario 1 (Pandex, also known as Engagement E1) and Scenario 2 (Bovia, or Engagement E2). The processing includes ingesting provenance graph data into a graph database as well as additional data integration and deduplication steps (Figure \ref{Fig:ALADAEN}, step 1). The final data includes a number of Boolean-valued datasets
, with each representing an aspect of the behavior of system processes (Figure \ref{Fig:ALADAEN}, step 2). Each row in such a dataset is a data point representing a single process run on the respective OS. It is expressed as a Boolean vector whereby a value of 1 in a vector cell indicates the corresponding attribute applies to that process. For instance, in Figure \ref{Fig:ALADAEN} (step 2, left subfigure) the process with id\\ \verb|ee27fff2-a0fd-1f516db3d35f| has the following sequence of events: \verb|</usr/sbin/avahi-autoipd|, \verb|216.73.87.152|, \\\verb|EVENT_OPEN, EVENT_CONNECT, ...>|. Specifically, the relevant datasets are interpreted as follows:

\begin{itemize}
    \item \verb|ProcessEvent| (PE): Its attributes are event types performed by the processes. A value of 1 in \verb|process[i]| means the process has performed at least one event of type $i$.
    \item \verb|ProcessExec| (PX): The attributes are executable names that are used to start the processes.
    \item \verb|ProcessParent| (PP): Its attributes are executable names that are used to start the parents of the processes.
    \item \verb|ProcessNetflow| (PN): The attributes here represent IP addresses and port names that have been accessed by the processes.
    \item \verb|ProcessAll| (PA): This dataset is described by the disjoint union of all attribute sets from the previous datasets.
\end{itemize}
Overall, with two attack scenarios (Pandex, Bovia), four OS (BSD, Windows, Linux, Android) and five aspects (PE, PX, PP, PN, PA), a total of forty individual datasets are composed, as illustrated in Figure \ref{Fig:ALADAEN} (step 2, right subfigure). 
They are described in Table~\ref{datatable} whereby the last column provides the number of attacks in each dataset. The substantially imbalanced nature of the datasets is clearly seen here.
\begin{table*}
\centering
\small
\resizebox{0.99\textwidth}{!}{
\begin{tabular}{|l|l||l|l|l|l|l|l|l|l|}
\hline & Scenario & Size& $PE$   & $PX$  & $PP$  & $PN$     & $PA$  & $nb\_attacks$    & $\%\frac{nb\_attacks}{nb\_processes}$     \\ \hline \hline
BSD    & 1 &288 MB &76903 / 29  & 76698 / 107  & 76455 / 24  & 31 / 136  & 76903 / 296 & 13&0.02\\  
    & 2 &1.27 GB &224624 / 31  &224246 / 135  & 223780 / 37  & 42888 / 62 &  224624 / 265      & 11&0.004\\ \hline
Windows & 1 &743 MB & 17569 / 22    &  17552 / 215  &   14007 / 77        &   92 / 13963      & 17569 / 14431& 8&0.04\\  
   & 2 &9.53 GB& 11151 / 30    &  11077 / 388  & 10922 / 84  & 329 / 125      &  11151 / 606    &8&0.07\\ \hline
Linux  & 1 &2858 MB &247160 / 24 & 186726 / 154 & 173211 / 40 & 3125 / 81 & 247160 / 299  &25&0.01\\
    & 2 &25.9 GB &282087 / 25 & 271088 / 140 & 263730 / 45 &6589 / 6225 &  282104 / 6435      &46&0.01\\ \hline
Android& 1 &2688 MB&102 / 21     &102 / 42&0 / 0&8 / 17& 102 / 80&9&8.8\\
&2 &10.9 GB&12106 / 27     &12106 / 44&0 / 0&4550 / 213&12106 / 295 &13&0.10\\ \hline
\end{tabular}
}

\caption{Experimental datasets of DARPA's TC program used in our study. A dataset entry (columns 4 to 8) is described by a number of rows (processes) / number of columns (attributes). For instance, with ProcessAll (PA) obtained from the second scenario using Linux, the dataset has 282104 rows and 6435 attributes with 46 APT attacks (0.01\%) \cite{Benabderrahmane21}. }
 \label{datatable}
\end{table*}

\subsection{Experimental Setup and Reproducibility}
To ensure reproducibility, we summarize below the experimental 
setup used to train and evaluate ALADAEN.

\subsubsection{Hyperparameters:}
Both AutoEncoders (AE1, AE2) and the discriminator were trained with the 
Adam optimizer ($\beta_1=0.5, \beta_2=0.999$) using a learning rate of $10^{-4}$ 
and batch size of 128. Each encoder projects the input to a latent dimension 
of $k=32$, followed by a dropout layer with rate $p=0.2$. 
Training was performed for up to 300 epochs with early stopping 
(patience = 10 epochs) based on validation reconstruction loss. 
The dual AutoEncoder reconstruction loss was weighted by 
$\alpha=0.5$ and the adversarial loss contribution by $\lambda=0.5$. 
A fixed random seed was used for all runs to ensure reproducibility.

\subsubsection{Dataset Splits:}
For each DARPA dataset, benign processes were used to form the training pool 
for the AutoEncoders, while validation and test sets were constructed from 
disjoint subsets of both benign and anomalous processes. 
A fixed random seed was used for partitioning to ensure reproducibility.

\subsubsection{Active Learning Protocol:}
ALADAEN starts from a cold-start setting with only 20\% labeled normal 
samples iteratively queries a small batch of samples (by percentile) until the labeling budget is exhausted. Queried samples are incorporated into the labeled pool 
and the model is retrained incrementally.


\subsubsection{Computational environment}
The experiments were conducted on a machine running macOS 14.5 with an Apple M1 Max chip, 64 GB RAM. The anomaly detection models were tested on every one of the 40 datasets.
\subsection{Results}

\subsubsection{\textbf{Comparison with existing methods}:}
Table \ref{ndcgscoresevalAll} presents a performance comparison of the different anomaly detection algorithms across multiple operating systems (BSD, Windows, Linux, and Android) and attack scenarios (Pandex and Bovia). In addition to the table presenting the different nDCG scores for the anomaly detection methods, we also provide a heatmap in Figure \ref{Fig:ndcgoutput} to facilitate easy visual interpretation of the results. The rows in the heatmaps represent the anomaly detection methods, while the columns represent the datasets. The sub-figures on the left-hand side correspond to the first attack scenario, and the sub-figures on the right-hand side correspond to the second attack scenario. Empty cells in the heatmaps indicate cases where the algorithms did not finish processing. \\
Next, we provide a detailed performance comparison across different operating systems and attack scenarios, highlighting how each anomaly detection method performs in varying environments. This breakdown allows for a deeper understanding of the strengths and weaknesses of each approach in specific contexts.

\begin{itemize}
    \item BSD (Cadets):
    \begin{itemize}
        \item In the Pandex scenario, ALADAEN demonstrates superior performance with the highest nDCG score of 0.88 in the PA dataset, outperforming traditional methods such as AVF (0.52) and OC3 (0.38), VF-ARM (0.67 in PP) and VR-ARM (0.64 in PE). MAE-AD and PGAE-APT show moderate results (0.67 and 0.31 respectively), while DeepOCCL and RT-APT trail behind (0.22 and 0.36 in PP). ALADAEN's ability to effectively prioritize anomalies is further shown in the PX and PE configurations, where it scores 0.78 and 0.75 respectively, significantly better than the other approaches.
        \item For the Bovia scenario, ALADAEN continues to lead with an impressive nDCG score of 0.98 in the PP configuration, marking the highest score among all methods in this attack scenario. Notably, other methods like AVF, OC3, and FPOF either fail to complete (DNF) or produce much lower scores. This stark contrast highlights ALADAEN's robustness in complex, real-world attack scenarios.
    \end{itemize}

\item Windows (5dir):
\begin{itemize}
    \item In the Windows Pandex scenario, ALADAEN again stands out with an nDCG of 0.82 in the PE configuration, surpassing other methods like AVF and OC3. Additionally, VR-ARM also performs in this case with a high nDCG of 0.82 in the PE configuration, surpassing HDAPT-IDS (0.67 PE) and MAE-AD (0.77 PN). 
DeepOCCL and RT-APT achieve 0.57 and 0.67 (PN) respectively but remain below ALADAEN. 
    \item For the Bovia scenario, ALADAEN maintains a competitive position with a score of 0.45 in the PA configuration, outperforming methods like VR-ARM (0.35) and OC3 (0.24). Notably, several traditional methods, such as OD and FPOF, fail to provide any valid results (DNF) across most configurations, indicating their struggle with this scenario.
\end{itemize}

\item Linux (Trace):

\begin{itemize}
    \item In the Linux Pandex scenario, ALADAEN achieves the highest nDCG score of
0.77 in the PA configuration, clearly outperforming HDAPT-IDS (0.55) and VR-ARM
(0.58). ALADAEN also shows competitive performance in other configurations
like PE (0.65), highlighting its versatility across different data distributions. \item For the Bovia scenario, ALADAEN reaches a high score of 0.87 (PA), while RT-APT achieves the best score 0.97 (PN), suggesting that RT-APT performs strongly for network-centric behaviors, whereas ALADAEN remains consistently superior in multi-view settings. It shows competitive performance in other configurations like PN (0.77), highlighting its versatility across different data distributions.
\end{itemize}

\item Android (Clearscope):
\begin{itemize}
    \item In the Android Pandex scenario, ALADAEN performs exceptionally well, achieving an nDCG score of 0.867 in the PE configuration, slightly ahead of VR-ARM (0.87), closely followed by MAE-AD (0.85 PX) and PGAE-APT (0.82 PN). ALADAEN's strong performance in this scenario underlines its adaptability in mobile and embedded systems like Android.
\item For the Bovia scenario, ALADAEN maintains good performance with a score of 0.67 in PE, showing that it remains competitive even in more challenging scenarios. VR-ARM method reaches 0.71 in PN, surpassing ALADAEN in this case.\\
\end{itemize}

\end{itemize}

\subsubsection{\textbf{Key Insights}:}
\begin{figure}
     \centering
     \includegraphics[width=\linewidth]{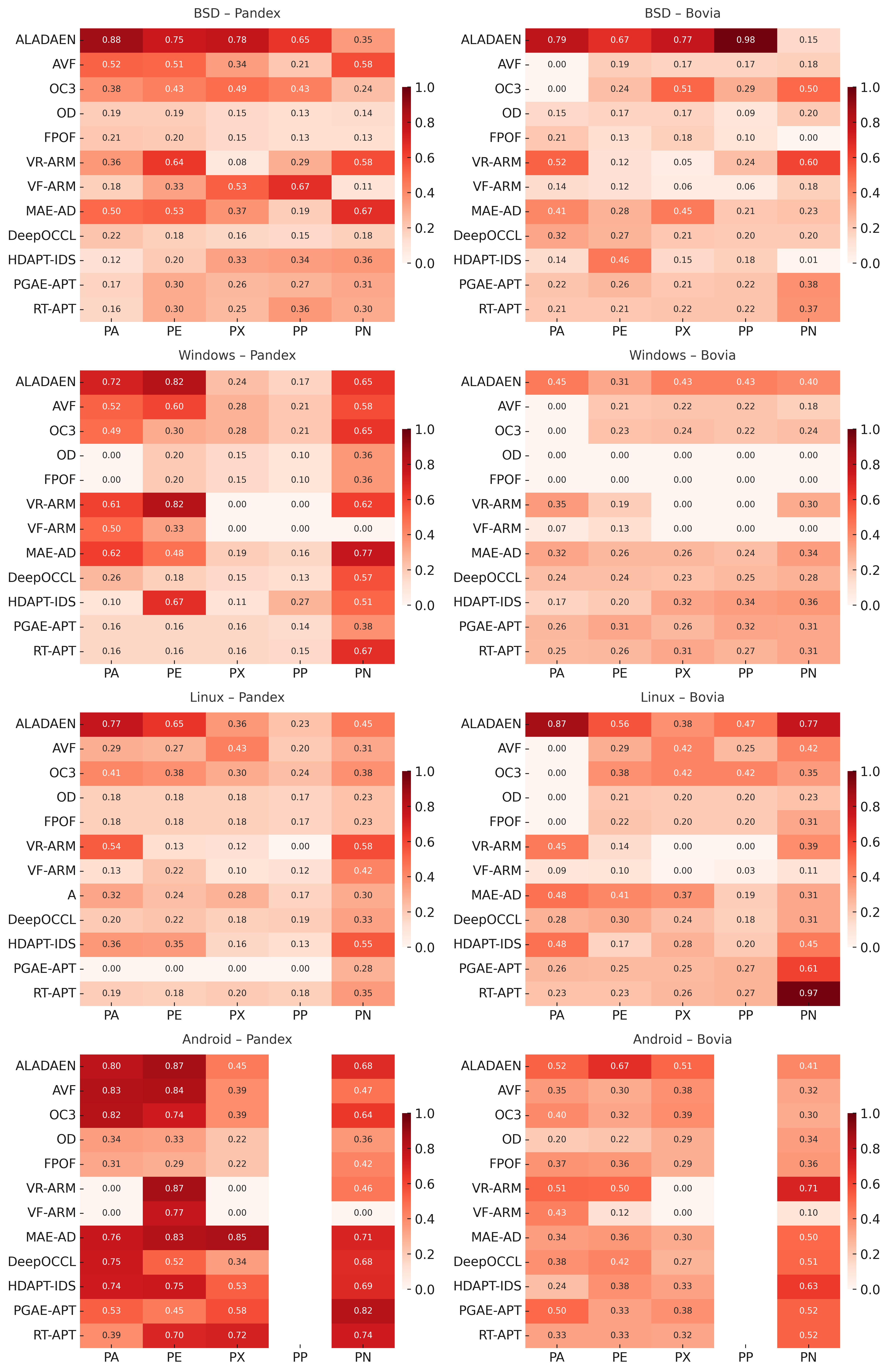}
     \caption{nDCG Score Comparison of Anomaly Detection Algorithms Across Operating Systems and Attack Scenarios. The rows represent the anomaly detection methods, while the columns represent the datasets. The subfigures on the left-hand side correspond to the first attack scenario, and the subfigures on the right-hand side correspond to the second attack scenario. }\label{Fig:ndcgoutput}
 
\end{figure}

As shown in Figure~\ref{Fig:ndcgoutput}, ALADAEN consistently achieves the highest or near-highest nDCG scores across all operating systems and attack scenarios, demonstrating its ability to generalize across diverse environments and data distributions. 
In high-complexity attack cases such as BSD–Bovia and Linux–Pandex, ALADAEN not only surpasses classical baselines but also outperforms the most recent deep learning models (MAE-AD, RT-APT, DeepOCCL, PGAE-APT, and HDAPT-IDS). In several cases, ALADAEN outperforms its closest competitors by a significant margin. For example, in BSD Bovia (PP configuration), ALADAEN achieves a score of 0.98, while OC3, AVF, and FPOF fail to reach 0.50. ALADAEN’s attention-driven adversarial dual-AutoEncoder design provides stable ranking even in extremely imbalanced datasets, where APT samples constitute less than 0.01\% of total processes. 
By contrast, methods like DeepOCCL and HDAPT-IDS exhibit high variance across configurations, struggling to maintain performance under cold-start or low-label regimes.\\
The integration of the adversarial loss stabilizes the latent representations, allowing ALADAEN to retain high nDCG scores even when the feature distributions differ across operating systems. 
This contrasts with MAE-AD, which shows stronger results on Android (mobile traces) but lower generalization to BSD and Linux server logs.

VR-ARM emerges as a strong competitor to ALADAEN, particularly in the Windows and Android environments, where it occasionally matches ALADAEN in specific configurations. For instance, VR-ARM achieves an nDCG of 0.82 in Windows $\times$ Pandex $\times$ PE, and 0.87 in Android $\times$ Pandex $\times$ PE forensic situations.\\
While AVF shows decent performance in some simpler scenarios (e.g., 0.84 in Android Pandex), it generally lags behind in more complex environments.\\
OC3 shows good performance in a few configurations, such as Android Pandex (PA), where it achieves a maximum score of 0.82. However, like AVF, OC3 struggles with complex scenarios, often producing DNF results, especially in higher-dimensional datasets.\\
OD (Outlier Degree) consistently ranks among the lowest-performing methods, with multiple DNF results across all operating systems and attack scenarios. This indicates that OD is not well-suited for detecting advanced threats in complex, high-dimensional environments.\\
Similarly, FPOF (Frequent Pattern Outlier Factor) struggles across most scenarios, failing to provide valid results in several configurations and consistently underperforming compared to ALADAEN and other methods.\\
Sometimes FPOF, OC3, and OD failed to produce results within a reasonable runtime, not terminating even after more than four hours on high-dimensional datasets. Such prolonged runtimes are undesirable in practical cybersecurity settings, where rapid anomaly detection and response are paramount to minimizing attack impact. In contrast, ALADAEN completed inference and ranking for all datasets within minutes, demonstrating not only higher detection accuracy but also superior suitability for time-sensitive operational environments.

The extended experiments reveal also distinct behavioral trends among the deep learning-based baselines. MAE-AD achieves competitive results in Android and BSD environments (e.g., 0.83 in Android–Pandex and 0.50 in BSD–Pandex), confirming the effectiveness of its masked reconstruction strategy in learning local feature dependencies. However, its reliance on context masking limits its ability to capture fine-grained, cross-view irregularities, leading to lower scores in Linux and Windows traces where temporal and causal diversity is higher. DeepOCCL demonstrates stable yet moderate performance across all datasets, reflecting its robust latent one-class modeling but also its sensitivity to high-dimensional feature noise in provenance graphs. HDAPT-IDS performs relatively well on Windows (0.67 in PE) and Android (0.75 in PE), leveraging hierarchical clustering to capture coarse attack patterns. Nonetheless, its multi-stage structure incurs computational overhead and fails to adapt rapidly under streaming data, causing inconsistent results in more dynamic domains such as Linux–Bovia. PGAE-APT, designed for graph-structured APT traces, produces strong outcomes in a few network-related views (e.g., PN on Android–Pandex = 0.82) but suffers performance degradation when applied to heterogeneous process-action data, highlighting its limited robustness outside graph formalisms. 
Finally, RT-APT stands out in network-intensive configurations, such as Linux–Bovia (0.97 in PN), validating the strength of its transformer-based temporal adaptivity. However, it underperforms in persistence-heavy, multi-phase contexts where reconstruction stability and balanced learning become crucial. 

Overall, these results confirm that while each recent method excels within its architectural niche, none achieves ALADAEN’s uniform superiority across all operating systems and views. 
ALADAEN’s integration of attention-driven dual adversarial autoencoding, (and later GAN-based augmentation, and active learning) yields a uniquely stable ranking capability, allowing it to outperform both reconstruction-oriented (MAE-AD, PGAE-APT) and threshold-adaptive (RT-APT) baselines across diverse provenance modalities.

Table \ref{tab:winning-algo} summarizes these findings, emphasizing the highest nDCG scores and the corresponding winner methods. For each OS/attack scenario, the maximum nDCG value is highlighted in red. Observe that ALADAEN performed best among the competitors in 6 out of 8 forensic configurations and demonstrated superior performance, achieving the highest scores in multiple datasets and scenarios.

\begin{table}
\tiny
\centering
\begin{tabular}{lccccccc}
\hline

\textbf{Operating System}             & \textbf{Attack Scenario} & \textbf{Algorithm} & \textbf{PA}                & \textbf{PE}     & \textbf{PX}      & \textbf{PP}  & \textbf{PN}   \\ \hline
\multirow{9}{*}{Cadets (BSD)}         & \multirow{7}{*}{Pandex}  & ALADAEN            & \cellcolor{red}\textbf{0.88}                      & 0.75                       & 0.78                       & 0.65                    & 0.35                      \\
                                      &                          & AVF                & 0.52                               & 0.51                      & 0.34                      & 0.21                    & 0.58                      \\
                                      &                          & OC3                & 0.38                               & 0.43                      & 0.49                      & 0.43                    & 0.24                      \\
                                      &                          & OD                 & 0.19                               & 0.19                      & 0.15                      & 0.13                    & 0.14                      \\
                                      &                          & FPOF               & 0.21                               & 0.2                       & 0.15                      & 0.13                    & 0.13                      \\
                                      &                          & VR-ARM             & 0.36                               & 0.64                      & 0.08                      & 0.29                    & 0.58                      \\
                                      &                          & VF-ARM             & 0.18                               & 0.33                      & 0.53                      & 0.67                    & 0.11         \\ 
                                      &&MAE-AD&0.5	&0.53&	0.37	&0.19	&0.67\\ &&DeepOCCL&0.22	&0.18	&0.16	&0.15	&0.18\\ 
                                      &&HDAPT-IDS&0.12	&0.2	&0.33	&0.34	&0.36\\
                                       &&PGAE-APT&0.17&	0.3	&0.26	&0.27	&0.31 \\
                                      \
                                       &&RT-APT&0.16&	0.3&	0.25	&0.36	&0.3 \\
                                       \cline{2-8}
                                      & \multirow{7}{*}{Bovia}   & ALADAEN            & 0.79                                & 0.67                      & 0.77                      & \cellcolor{red}\textbf{0.98}           & 0.15                      \\
                                      &                          & AVF                & DNF                                & 0.19                      & 0.17                      & 0.17                    & 0.18                       \\
                                      &                          & OC3                & DNF                               & 0.24                      & 0.51                      & 0.29                    & 0.50                       \\
                                      &                          & OD                 & 0.15                               & 0.17                      & 0.17                      & 0.09                    & 0.20                      \\
                                      &                          & FPOF               & 0.21                               & 0.13                      & 0.18                      & 0.1                     & DNF                       \\
                                      &                          & VR-ARM             & 0.52                               & 0.12                      & 0.05                      & 0.24                    & 0.6                       \\
                                      &                          & VF-ARM             & 0.14                               & 0.12                      & 0.06                      & 0.06                    & 0.18         \\ 
                                      &&MAE-AD&0.41	&0.28	&0.45	&0.21	&0.23\\ &&DeepOCCL&0.32	&0.27	&0.21	&0.2	&0.2 \\ 
                                      &&HDAPT-IDS&0.14	&0.46	&0.15	&0.18	&0.01 \\
                                       &&PGAE-APT&0.22&	0.26	&0.21	&0.22	&0.38\\
                                      \
                                       &&RT-APT&0.21&	0.21	&0.22	&0.22	&0.37 \\ \hline
\multirow{8}{*}{5dir (Windows)}       & \multirow{7}{*}{Pandex}  & ALADAEN            & 0.72                               & \cellcolor{red}\textbf{0.82}              & 0.24                      & 0.17                    & 0.65                       \\
                                      &                          & AVF                & 0.52                               & 0.6                       & 0.28                      & 0.21                    & 0.58                      \\
                                      &                          & OC3                & 0.49                                & 0.3                       & 0.28                      & 0.21                    & 0.65                      \\
                                      &                          & OD                 & DNF                                & 0.2                       & 0.15                      & 0.1                     & 0.36                      \\
                                      &                          & FPOF               & DNF                                & 0.2                       & 0.15                      & 0.1                     & 0.36                      \\
                                      &                          & VR-ARM             & 0.61                               & \cellcolor{red}\textbf{0.82}             & 0                         & 0                       & 0.62                      \\
                                      &                          & VF-ARM             & 0.5                                & 0.33                      & 0                         & 0                       & 0           \\ 
                                      &&MAE-AD&0.62	&0.48	&0.19	&0.16	&0.77 \\ &&DeepOCCL&0.26	&0.18	&0.15	&0.13	&0.57              \\ 
                                      &&HDAPT-IDS&0.1	&0.67	&0.11	&0.27	&0.51\\
                                       &&PGAE-APT&0.16&	0.16	&0.16	&0.14	&0.38 \\
                                      \
                                       &&RT-APT&0.16&	0.16	&0.16	&0.15	&0.67 \\ \cline{2-8} 
                                      & \multirow{7}{*}{Bovia}   & ALADAEN            & \cellcolor{red}\textbf{0.45}                      & 0.31                      & 0.43                      & 0.43                    & 0.4                       \\
                                      &                          & AVF                & DNF                                & 0.21                      & 0.22                      & 0.22                    & 0.18                       \\
                                      &                          & OC3                & DNF                                & 0.23                      & 0.24                      & 0.22                    & 0.24                       \\
                                      &                          & OD                 & DNF                                & DNF                       & DNF                       & DNF                     & DNF                       \\
                                      &                          & FPOF               & DNF                                & DNF                       & DNF                       & DNF                     & DNF                       \\
                                      &                          & VR-ARM             & 0.35                               & 0.19                      & 0                         & 0                       & 0.3                       \\
                                      &                          & VF-ARM             & 0.07                               & 0.13                      & 0                         & 0                       & 0           \\ 
                                      &&MAE-AD&0.32	&0.26	&0.26	&0.24	&0.34\\ &&DeepOCCL&0.24	&0.24	&0.23	&0.25	&0.28\\ 
                                      &&HDAPT-IDS&0.17	&0.2	&0.32	&0.34	&0.36 \\
                                       &&PGAE-APT&0.26	&0.31	&0.26	&0.32	&0.31 \\
                                      \
                                       &&RT-APT&0.25	&0.26	&0.31	&0.27	&0.31 \\ \hline
\multirow{8}{*}{Trace (Linux)}        & \multirow{7}{*}{Pandex}  & ALADAEN            & \cellcolor{red}\textbf{0.77}                      & 0.65                       & 0.36                      & 0.23                    & 0.45                       \\
                                      &                          & AVF                & 0.29                               & 0.27                      & 0.43                      & 0.2                     & 0.31                      \\
                                      &                          & OC3                & 0.41                                & 0.38                      & 0.3                       & 0.24                    & 0.38                      \\
                                      &                          & OD                 & 0.18                                & 0.18                      & 0.18                      & 0.17                    & 0.23                      \\
                                      &                          & FPOF               & 0.18                                & 0.18                      & 0.18                      & 0.17                    & 0.23                      \\
                                      &                          & VR-ARM             & 0.54                               & 0.13                      & 0.12                      & 0                       & 0.58                      \\
                                      &                          & VF-ARM             & 0.13                               & 0.22                      & 0.1                       & 0.12                    & 0.42       \\ 
                                      &&MAE-AD&0.32	&0.24	&0.28	&0.17	&0.3 \\ &&DeepOCCL&0.2	&0.22&	0.18	&0.19	&0.33 \\ 
                                      &&HDAPT-IDS&0.36	&0.35	&0.16	&0.13	&0.55 \\
                                       &&PGAE-APT&DNF	&DNF	&DNF	&DNF	&0.28 \\
                                      \
                                       &&RT-APT&0.19&	0.18	&0.2	&0.18	&0.35 \\ \cline{2-8} 
                                      & \multirow{7}{*}{Bovia}   & ALADAEN            & 0.87                     & 0.56                      & 0.38                      & 0.47                    & 0.77                      \\
                                      &                          & AVF                & DNF                                & 0.29                      & 0.42                      & 0.25                    & 0.42                       \\
                                      &                          & OC3                & DNF                               & 0.38                      & 0.42                      & 0.42                    & 0.35                       \\
                                      &                          & OD                 & DNF                                & 0.21                      & 0.2                       & 0.2                     & 0.23                       \\
                                      &                          & FPOF               & DNF                                & 0.22                      & 0.2                       & 0.2                     & 0.31                       \\
                                      &                          & VR-ARM             & 0.45                               & 0.14                      & 0                         & 0                       & 0.39                      \\
                                      &                          & VF-ARM             & 0.09                               & 0.1                       & 0.004                     & 0.03                    & 0.11        \\ 
                                      &&MAE-AD&0.48&	0.41	&0.37	&0.19	&0.31 \\ &&DeepOCCL&0.28	&0.3	&0.24	&0.18	&0.31              \\ 
                                      &&HDAPT-IDS&0.48	&0.17	&0.28	&0.2	&0.45 \\
                                       &&PGAE-APT&0.26&	0.25	&0.25	&0.27	&0.61 \\
                                      \
                                       &&RT-APT&0.23&	0.23	&0.26	&0.27	&\cellcolor{red}\textbf{0.97} \\ \hline
\multirow{8}{*}{Clearscope (Android)} & \multirow{7}{*}{Pandex}  & ALADAEN            & 0.80                               & \cellcolor{red}\textbf{0.87 }           & 0.45                      & NA                      & 0.68                      \\
                                      &                          & AVF                & 0.83                               & 0.84                      & 0.39                      & NA                      & 0.47                      \\
                                      &                          & OC3                & 0.82                              & 0.74                      & 0.39                      & NA                      & 0.64                      \\
                                      &                          & OD                 & 0.34                               & 0.33                      & 0.22                      & NA                      & 0.36                      \\
                                      &                          & FPOF               & 0.31                               & 0.29                      & 0.22                      & NA                      & 0.42                      \\
                                      &                          & VR-ARM             & 0                                  & \cellcolor{red}\textbf{0.87}                      & 0                         & NA                      & 0.46                      \\
                                      &                          & VF-ARM             & 0                                  & 0.77                      & 0                         & NA                      & 0          \\ 
                                      &&MAE-AD&0.76&	0.83	&0.85	&NA	&0.71 \\ &&DeepOCCL&0.75	&0.52	&0.34	&NA&	0.68\\ 
                                      &&HDAPT-IDS&0.74	&0.75	&0.53	&NA	&0.69 \\
                                       &&PGAE-APT&0.53&	0.45	&0.58&	NA	&0.82 \\
                                      \
                                       &&RT-APT&0.39&	0.7	&0.72	&NA	&0.74 \\ \cline{2-8} 
                                      & \multirow{7}{*}{Bovia}   & ALADAEN            & \multicolumn{1}{c}{0.52}          & \multicolumn{1}{c}{0.67} & \multicolumn{1}{c}{0.51} & \multicolumn{1}{c}{NA} & \multicolumn{1}{c}{0.41} \\
                                      &                          & AVF                & \multicolumn{1}{c}{0.35}          & \multicolumn{1}{c}{0.3}  & \multicolumn{1}{c}{0.38} & \multicolumn{1}{c}{NA} & \multicolumn{1}{c}{0.32} \\
                                      &                          & OC3                & \multicolumn{1}{c}{0.40} & \multicolumn{1}{c}{0.32} & \multicolumn{1}{c}{0.39} & \multicolumn{1}{c}{NA} & \multicolumn{1}{c}{0.30}  \\
                                      &                          & OD                 & \multicolumn{1}{c}{0.20}          & \multicolumn{1}{c}{0.22} & \multicolumn{1}{c}{0.29} & \multicolumn{1}{c}{NA} & \multicolumn{1}{c}{0.34}  \\
                                      &                          & FPOF               & \multicolumn{1}{c}{0.37}          & \multicolumn{1}{c}{0.36} & \multicolumn{1}{c}{0.29} & \multicolumn{1}{c}{NA} & \multicolumn{1}{c}{0.36} \\
                                      &                          & VR-ARM             & \multicolumn{1}{c}{0.51}          & \multicolumn{1}{c}{0.5}  & \multicolumn{1}{c}{0}    & \multicolumn{1}{c}{NA} & \multicolumn{1}{c}{\cellcolor{red}\textbf{0.71}} \\
                                      &                          & VF-ARM             & \multicolumn{1}{c}{0.43}          & \multicolumn{1}{c}{0.12} & \multicolumn{1}{c}{0}    & \multicolumn{1}{c}{NA} & \multicolumn{1}{c}{0.1}  \\ 
                                      &&MAE-AD&0.34	&0.36	&0.3	&NA	&0.5\\ &&DeepOCCL&0.38	&0.42	&0.27	&NA&	0.51\\ 
                                      &&HDAPT-IDS&0.24	&0.38	&0.33	&NA	&0.63\\
                                       &&PGAE-APT&0.5	&0.33	&0.38	&NA	&0.52\\
                                      \
                                       &&RT-APT&0.33&	0.33	&0.32&	NA	&0.52 \\ \hline
\end{tabular}

\caption{Performance Evaluation of Anomaly Detection Methods Using nDCG Scores. Bold values represent the max nDCG for each forensic configuration (OS $\times$ attack scenario $\times$ dataset). NA: data not available. DNF: Did not finish.}
\label{ndcgscoresevalAll}
\end{table}

\begin{table*}
\centering
\Rotatebox{90}{
\small
{%
\begin{tabular}{lllccccc}
\hline
\textbf{OS}&\textbf{Scenario} &\textbf{  }                                                                                                             & 
\textbf{PA} & \textbf{PE} & \textbf{PX} & \textbf{PP} & \textbf{PN} \\ \hline
\multirow{4}{*}{\textbf{Cadets (BSD)}}         & \multirow{2}{*}{\textbf{Pandex}} & Highest nDCG   & \cellcolor{red}\textbf{0.88 }      & 0.75         & 0.78        & 0.67          & 0.58            \\
                                               &                                  &Winning method  & \cellcolor{red}ALADAEN           & ALADAEN            & ALADAEN            & VR-ARM             & VR-ARM             \\ \cline{2-8} 
                                               & \multirow{2}{*}{\textbf{Bovia}}  & Highest nDCG   & 0.79       & 0.67         & 0.77        & \cellcolor{red}\textbf{0.98}          & 0.60           \\
                                               &                                  & Winning method & ALADAEN          & ALADAEN            & ALADAEN           & \cellcolor{red}ALADAEN             & VR-ARM               \\ \hline
\multirow{4}{*}{\textbf{5dir (Windows)}}       & \multirow{2}{*}{\textbf{Pandex}} & Highest nDCG   & 0.72       & \cellcolor{red}\textbf{0.82}         & 0.28        & 0.27           & 0.77             \\
                                               &                                  & Winning method & ALADAEN          & \cellcolor{red}VR-ARM, ALADAEN            & AVF,OC3           & HDAPT-IDS        & MAE-AD              \\ \cline{2-8} 
                                               & \multirow{2}{*}{\textbf{Bovia}}  & Highest nDCG   & \cellcolor{red}0.45       & 0.31         & 0.43        & 0.43         & 0.40           \\
                                               &                                  & Winning method & \cellcolor{red}ALADAEN           & ALADAEN             & ALADAEN            &  ALADAEN              & ALADAEN              \\ \hline
\multirow{4}{*}{\textbf{Trace (Linux)}}        & \multirow{2}{*}{\textbf{Pandex}} & Highest nDCG   & \cellcolor{red}\textbf{0.77}       & 0.65         & 0.43          & 0.24            & 0.58            \\
                                               &                                  & Winning method & \cellcolor{red}ALADAEN          & ALADAEN            & AVF           & OC3             & VR-ARM             \\ \cline{2-8} 
                                               & \multirow{2}{*}{\textbf{Bovia}}  & Highest nDCG   & 0.87       & 0.56         & 0.42        & 0.47            & \cellcolor{red}\textbf{0.97}           \\
                                               &                                  & Winning method & ALADAEN           & ALADAEN             & AVF, OC3            & ALADAEN             & \cellcolor{red}RT-APT               \\ \hline
\multirow{4}{*}{\textbf{Clearscope (Android)}} & \multirow{2}{*}{\textbf{Pandex}} & Highest nDCG   & 0.83         &\cellcolor{red} \textbf{0.87}           & 0.72        & NA              & 0.68             \\
                                               &                                  & Winning method & AVF          & \cellcolor{red}VR-ARM, ALADAEN            & RT-APT           & NA              & ALADAEN              \\ \cline{2-8} 
                                               & \multirow{2}{*}{\textbf{Bovia}}  & Highest nDCG   & 0.52         & 0.67         & 0.51        & NA            & \cellcolor{red}\textbf{0.71}              \\
                                               &                                  & Winning method & ALADAEN          & ALADAEN             & ALADAEN            &NA             & \cellcolor{red}VR-ARM              \\ \hline
\end{tabular}
}
}
\caption{Highest nDCG scores achieved and the corresponding algorithms on all available datasets. Red highlighted values represent the winner methods with their best nDCG score, for each forensic configuration: OS $\times$ attack scenario $\times$ dataset.}
\label{tab:winning-algo}
\end{table*}

\subsubsection{\textbf{Active learning assessment}}
Next, we present the evaluation of ALADAEN when including the active learning loop. Given that we are working with forty databases, it is almost impractical to discuss and comment the nDCG scores variation for each single dataset. Hence, we will first present the results for the cases where ALADAEN achieved the best performances when compared with the previous anomaly detection approaches, as highlighted in Table \ref{tab:winning-algo}. For the remaining datasets, a summary discussion will follow, providing insights into their behavior. This would allow us to focus on the most significant results, while still offering a comprehensive understanding of the model’s performance across all datasets.\\
For the active learning process implemented in ALADAEN, the parameter $N_{\text{iterations}}$ has been set to 40. This number was selected to balance the need for model improvement with the practical constraints of real-world cybersecurity applications. In the context of cyber threats, security experts must act swiftly to mitigate risks and prevent potential attacks. Prolonged iteration cycles could delay critical decision-making, leaving systems vulnerable during the learning phase. By limiting the iterations to 40, we ensure that the model reaches an optimal performance level without causing delays that could compromise security response times. This approach aligns with the urgency required in cybersecurity, where timely interventions are paramount. While other values for the number of iterations are possible, they should not be too prolonged, as extended cycles could hinder the swift decision-making required in cybersecurity scenarios.

\begin{figure}[!htb]
     \centering
     \includegraphics[width=\linewidth]{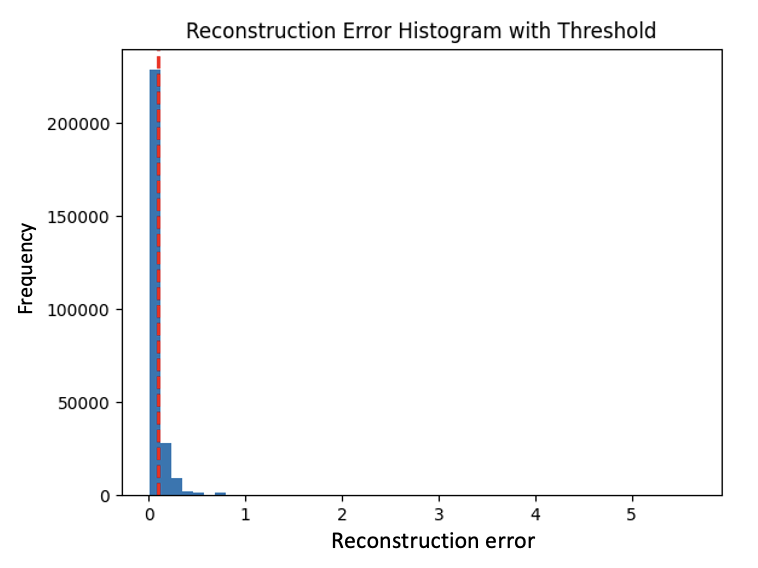}
     \caption{Reconstruction Error Histogram with Threshold, where the x-axis represents the reconstruction error values, and the y-axis indicates the frequency (count) of data points with a certain error.}\label{Fig:reconerror}
 
\end{figure}

\paragraph{Uncertainty Sample Selection:} We first examine the reconstruction error distribution and the selection strategy for uncertain samples near the decision threshold $\tau$. This analysis illustrates how ALADAEN identifies the most informative data points for oracle querying in each iteration, ensuring that the labeling budget is spent on samples that most improve the decision boundary.  
Figure \ref{Fig:reconerror} represents the reconstruction error histogram of the neural network with the threshold $\tau$ in red line, where the x-axis represents the reconstruction error values, and the y-axis indicates the frequency (count) of data points with a certain error. In this example data belongs to Linux$\times$PA$\times$Pandex forensic configuration (randomly chosen here for illustration purposes).
We can observe that most data points have low reconstruction errors, concentrated near zero. This suggests that ADAEN AutoEncoder is performing well on most of the input data, reconstructing it with high accuracy.\\
The red dashed line represents the threshold marker $\tau$ that is chosen to distinguish between normal and anomalous data points (since $\text{Anomaly}(x)=1 ~\text{if}~error(x)>\tau$). Points with reconstruction error above the threshold would be classified as anomalies, while those below it are considered normal. Here, the uncertainty threshold $\tau$ for anomaly detection is determined using a percentile-based approach. Specifically, the threshold $\tau$ is set at the 80th percentile of the reconstruction error values calculated on the test dataset. This means that 80\% of the data points are expected to have reconstruction errors below this threshold, and only the remaining 20\% will be considered potentially anomalous. This method ensures that the threshold adapts to the distribution of reconstruction errors, enabling more flexible and data-driven anomaly detection. It is important to note that the choice of the 80th percentile as the uncertainty threshold is flexible and can be adjusted to fine-tune the algorithm's performance. By selecting different percentile values, the threshold can be adapted to balance the trade-off between detecting anomalies and reducing false positives, allowing the algorithm to be tailored to the specific characteristics and requirements of the dataset or application scenario. Note that it is also possible for the model to update iteratively the threshold in an automatic manner. More specifically, the threshold could be recalibrated at each iteration of active learning, to a different percentile of the reconstruction error for normal instances. This allows the model to incorporate feedback and refine its detection sensitivity, improving its ability to distinguish between normal and anomalous data points over time.\\
This histogram is a useful way to visualize the performance of the AutoEncoder, as it shows how well the model distinguishes between normal and anomalous data based on reconstruction errors.
Data points with high reconstruction error might represent the anomalous points that the model struggles to reconstruct, signaling potential attacks or anomalies. The AutoEncoder in this Linux$\times$PA$\times$Pandex forensic configuration has learned to accurately reconstruct the majority of normal data (errors close to zero), and the threshold helps detect anomalies by flagging instances with higher reconstruction errors. Similar behaviors have also been observed with the other datasets.

 \begin{figure}[!htb]
     \centering
     \includegraphics[width=\linewidth]{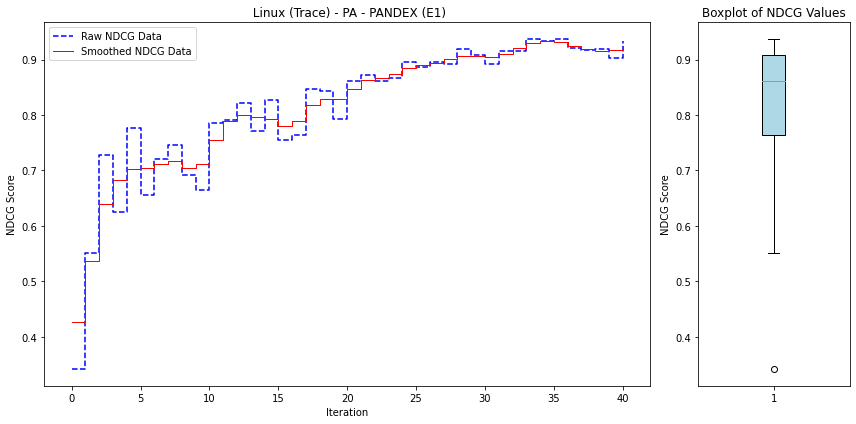}
     \caption{nDCG score variation over Active Learning iterations for Linux PA dataset using the ALADAEN framework (Pandex E1 scenario). Values in x-axis represent active learning iterations, whereas y-axis contains nDCG scores. The raw nDCG data is represented by the blue dashed line, while the smoothed nDCG values are depicted by the red dashed line. On the right, a boxplot of the nDCG values shows the distribution of scores throughout the Active Learning process. The figure highlights the model's performance improvements over iterations, stabilizing at higher nDCG scores as more data is incorporated.}\label{Fig:PEADANE}
 
\end{figure}

\paragraph{Ranking Improvement Across Iterations:} We next evaluate how active learning improves anomaly ranking over time.
Figure \ref{Fig:PEADANE} presents the evolution of the nDCG scores over several Active Learning iterations for the Linux$\times$PA$\times$Pandex forensic configuration. It consists of two subfigures: The raw nDCG data is represented by the blue dashed line showing the progression of the nDCG scores over 40 iterations, while the smoothed nDCG values are depicted by the red curve. Initially, the nDCG score starts around 0.34 (min value at cold start) but gradually improves as iterations progress, stabilizing between 0.70 and 0.94 in later iterations. The median is about 0.84 and the mean over all values is 0.82. There are noticeable fluctuations in performance, indicating varying success rates at different stages. However, the overall trend is upward, demonstrating that the model is learning and improving its ranking accuracy with each iteration. \\
The figure provides several important insights into the evolution of nDCG scores during active learning iterations for this Linux$\times$PA$\times$Pandex configuration. The nDCG raw values exhibit fluctuations, especially in the early iterations, where the scores rise quickly, indicating that the model is learning and improving. However, there are some sharp drops followed by recoveries, signaling instability or challenging data points for the model. The smoothed values are obtained using a Gaussian filter. The smoothing reduces the noise from the raw values, giving a clearer trend. The red curve shows a general upward trajectory, indicating that the model is improving its ranking performance over time. The smoothing helps highlight that, despite some volatility, the model consistently improves after a certain number of iterations ($\approx$15). After the initial low score, which is about 0.34, the detection rate quickly increases, showing the effectiveness of active learning and GANs in improving the model’s performance as it selects informative samples. During mid iterations, some fluctuations in the raw data suggest that the model might face difficult instances or noise, but the overall trend is still positive, with nDCG scores stabilizing around 0.8 after around 25 iterations.
In later iterations, the smoothed curve flattens, indicating that the model has likely reached a plateau where additional iterations lead to marginal improvements. Observing such a plateau is an indication that it might not be useful to acquire new data to train the model since the generalization performance of the model will not increase anymore. A major outcome of this experiment also highlights the effectiveness of combining active learning with GAN-based data augmentation in improving anomaly detection performance. In fact, the model achieved a maximum nDCG score of around 0.94, surpassing the results obtained without active learning which is equal to 0.77. This improvement can be attributed to two key factors: (1) Active learning enables the model to focus on the most uncertain samples, which are critical in refining the decision boundary between normal and anomalous data. By iteratively selecting and labeling these ambiguous points, the model becomes more confident in its classifications. (2) The GAN-generated synthetic data enriches the training set by introducing realistic, diverse samples that reflect the normal data distribution. This enhanced dataset allows the model to generalize better, reducing overfitting and improving its ability to distinguish between normal and anomalous data. Together, these techniques contribute to the model's improved performance, demonstrating their ability to boost detection accuracy in scenarios with limited labeled data.\\
The right Boxplot in the same figure summarizes the distribution of nDCG scores throughout the entire active learning process. The box (light blue) represents the Inter Quartile Range (IQR), which contains the middle of the nDCG scores.
The median is around 0.84, showing that half the nDCG scores are above this value.
The whiskers extend to show the range of most of the data. The lower whisker reaches down to about 0.34, representing the lowest scores from early iterations.
The majority of nDCG scores are above 0.75, indicating that the model performs well overall, with only a few low outliers. The presence of outliers suggests that the model faced difficulty in some specific iterations but these instances are not representative of the overall trend. This boxplot gives a concise overview of the model's performance, showing that the majority of the iterations achieved higher scores, but there were some underperforming iterations as well.\\

\begin{figure}
    \centering
    \includegraphics[width=1\linewidth]{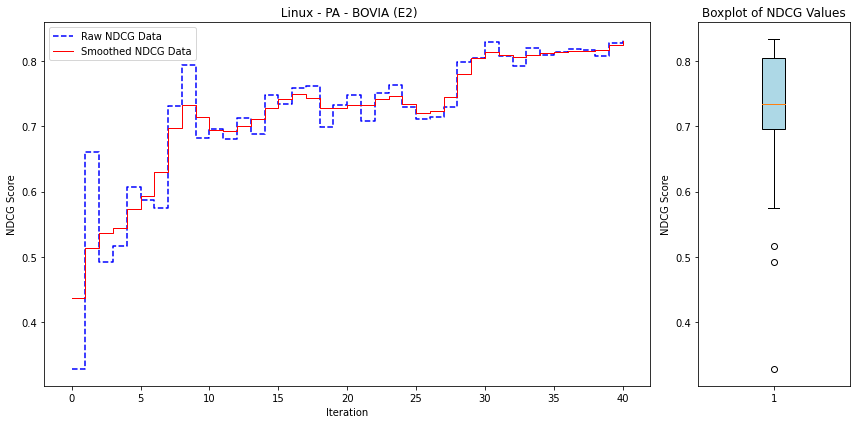}
    \caption{nDCG score variation over Active Learning iterations for the Linux PA dataset (Bovia E2 scenario).}
    \label{fig:LinuxPABovia}
\end{figure}

Similarly, in Figure \ref{fig:LinuxPABovia}, we observe the variation of nDCG scores during the active learning process for the Linux PA - Bovia (E2) scenario. The raw nDCG scores show fluctuations during the initial iterations, followed by a steady improvement as the active learning process progresses. The smoothed nDCG curve provides a clearer trend, illustrating a general upward trajectory, indicating that the model is learning and improving over time.\\
The boxplot also shows here a median score around 0.73, and the IQR is relatively narrow, suggesting consistency in the nDCG performance. A few lower outliers are observed, but the model generally performs well, with a peak nDCG score of approximately 0.84. Overall, this suggests that active learning is helping the model refine its ability to distinguish between normal and anomalous instances in this attack scenario. \\

\begin{figure}
    \centering
    \includegraphics[width=1\linewidth]{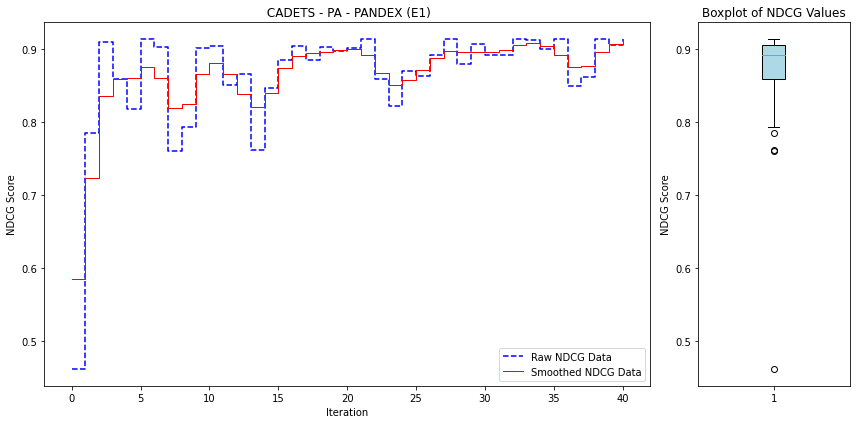}
    \caption{nDCG score variation over Active Learning iterations for the BSD (Cadets) PA dataset (Pandex E1 scenario).}
    \label{fig:CadersPAPandex}
\end{figure}

Figure \ref{fig:CadersPAPandex} shows the results across iterations for the BSD PA dataset with the Pandex (E1) attack scenario. 
The raw nDCG scores also show
here fluctuating behavior but generally improve over the iterations, reaching peaks above 0.9. The smoothed curve captures the trend more clearly, showing an upward trajectory and stabilizing close to 0.9 in the latter iterations, indicating improved performance over time.
The boxplot reveals a strong overall performance with a median around 0.89. The IRQ range is relatively tight, indicating stable performance. There are some lower outliers below 0.6, but most values are concentrated in the higher range, suggesting that the model has consistently good performance.\\
The next Figure \ref{fig:CadersPPBovia} presents the scores variation for the BSD PP dataset with Bovia (E2) attack scenario. The raw nDCG scores show a rapid rise to almost 1.0 by the 5th iteration, and remain consistently high throughout the subsequent iterations.
The smoothed data closely follows the raw data, indicating that the performance stabilizes around 1.0 after the initial iterations. The boxplot shows a highly concentrated nDCG score distribution near 1.0, with just one outlier below 0.70, indicating extremely high and consistent performance after the early stages of training. This figure indicates clearly that the model achieved near-optimal performance very early in the active learning process. After around 5 iterations, the nDCG score hovers around 1.0, demonstrating that the model effectively discriminates between normal and anomalous instances. The outlier in the boxplot likely corresponds to the initial phase of learning before the model stabilizes, but overall, the figure shows a robust and well-learned behavior under this attack scenario.\\
\begin{figure}
    \centering
    \includegraphics[width=1\linewidth]{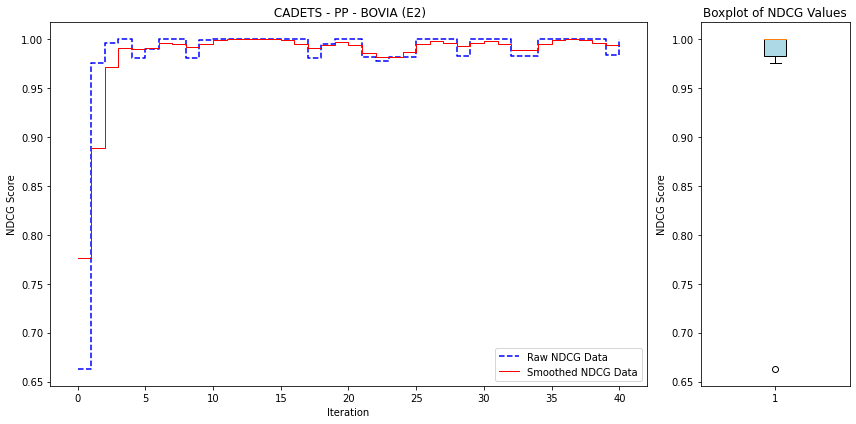}
    \caption{nDCG score variation over Active Learning iterations for the BSD (Cadets) PP dataset (Bovia E2 scenario).}
    \label{fig:CadersPPBovia}
\end{figure}
The next Figure \ref{fig:windowsPEPandex} shows the convergences for Windows PE Dataset in Pandex Attack Scenario. Here, the nDCG values start relatively high and quickly rise, indicating an initial improvement in the model's ranking performance within the first 10 iterations. After iteration 10, the performance fluctuates between 0.85 and 0.98, showing some variability in the ranking performance, which could be due to the attack scenario. The peak nDCG score approaches 1.0, which indicates the model is able to achieve optimal ranking performance during some iterations. The boxplot shows a median close to 0.98, suggesting that most iterations perform well. There are some outliers (below 0.80 and near 0.70), indicating occasional poor performance in certain iterations.
While the general performance is strong, the fluctuations in the raw nDCG data show that there are points where the model could be further optimized.\\
\begin{figure}
    \centering
    \includegraphics[width=1\linewidth]{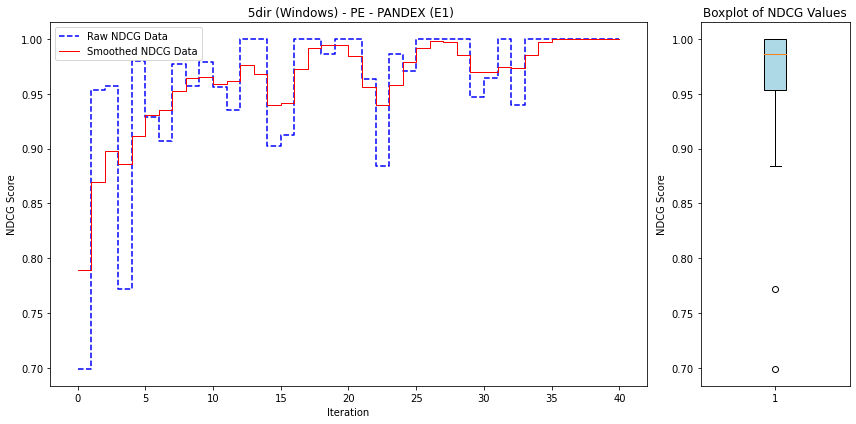}
    \caption{nDCG score variation over Active Learning iterations for the Windows PE dataset (Pandex E1 scenario).}
    \label{fig:windowsPEPandex}
\end{figure}
The graph in Figure \ref{fig:windowsPABovia} concerns 
Windows PA dataset (Bovia E2 scenario) and shows fluctuations in the nDCG score over 40 iterations, both in the raw and smoothed data. The raw data shows notable ups and downs, particularly in the early iterations (up to around iteration 10). Over time, the nDCG score improves and stabilizes at higher values, with some smaller variations around the 20th and 30th iterations. After the initial phase of fluctuation, the smoothed nDCG score begins to rise steadily from around iteration 10, reaching values close to 1, indicating high accuracy. The smoothed curve illustrates the large variations in the raw data, reflecting the underlying trend of performance improvement. Toward the final iterations, the score reaches 1, indicating good ranking quality. The boxplot also indicates a median nDCG score slightly below 0.9, with the IRQ range extending from about 0.75 to 0.95. The whiskers indicate that some values go as low as 0.4 and as high as 1. There are no extreme outliers, but the lower bound suggests that at certain points, performance can drop substantially. However, the concentration of the IQR near the higher end shows that overall ranking quality is generally quite good. Hence, the performance of the model in this configuration is relatively good, as indicated by the final stabilized nDCG scores nearing 1. However, early iterations and occasional fluctuations suggest that the model took time to stabilize, and there were occasional challenges in ranking quality during the process. This scenario demonstrates that while the model performs well overall, there are points where its ranking quality can degrade temporarily, likely due to variations in data handling during the attack scenario.\\
\begin{figure}[h]  
    \centering
    \includegraphics[width=1\linewidth]{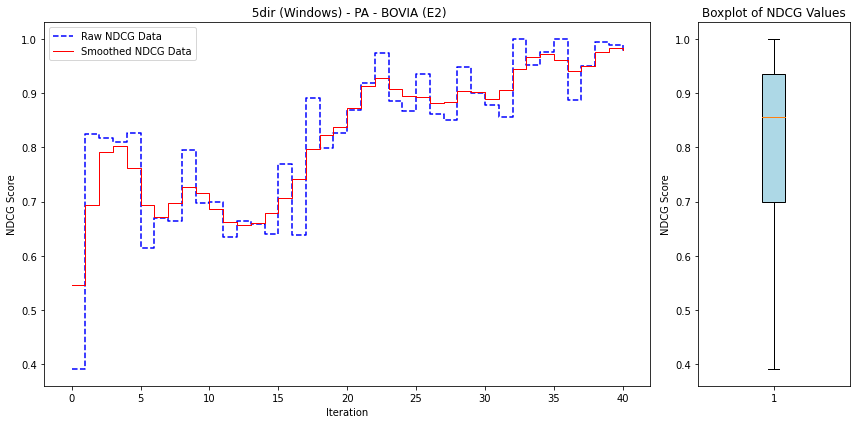}
    \caption{nDCG score variation over Active Learning iterations for the Windows PA dataset (Bovia E2 scenario).}
    \label{fig:windowsPABovia}
\end{figure}
Figure \ref{fig:androidPEPandex} contains scores for the Android PE dataset within Pandex (E1) attack scenario. Here, the nDCG score fluctuates between approximately 0.4 and 1.0 showing high variability with frequent jumps and drops in score across iterations. The smoothed values show a clearer upward progression over time. It generally increases and stabilizes around 0.9 after the initial fluctuations. Similarly, the boxplot shows that the median nDCG score appears to be close to 1.0, and the IQR suggests that most nDCG scores are high. There are also some outliers between 0.7 and 0.4. Albeit there is some initial volatility in performance over iterations, but this stabilizes as the iterations progress. The presence of outliers in the boxplot indicates occasional poor performance during cold start, but after that these are relatively infrequent compared to the high overall scores.\\
\begin{figure}[h]  
    \centering
    \includegraphics[width=1\linewidth]{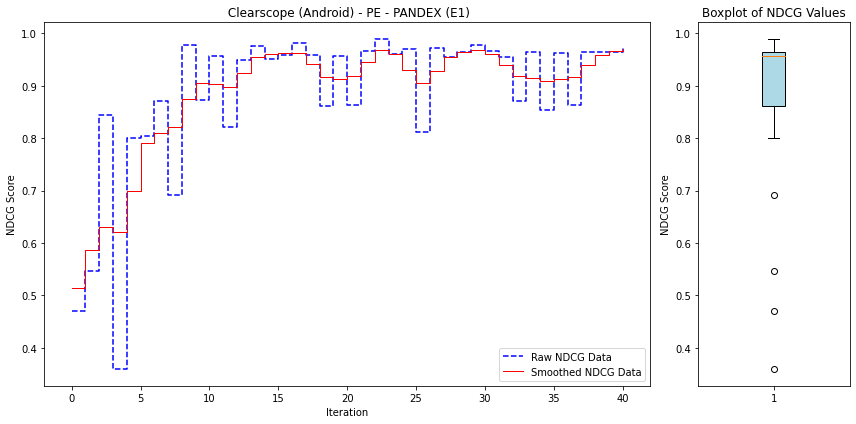}
    \caption{nDCG score variation over Active Learning iterations for the Android (Clearscope) PE dataset (Pandex E1 scenario).}
    \label{fig:androidPEPandex}
\end{figure}
Next, in Figure \ref{fig:androidPEBovia} the results are related to the Android-PE dataset within the Bovia (E2) scenario. There is noticeable variability, especially in the earlier iterations, where the score shows significant dips and peaks, indicating unstable performance during the beginning of the process. Smoother values show an increasing trend after 15th iteration, with scores gradually rising from around 0.7 to 0.8 before stabilizing and remaining relatively steady in the later iterations. The boxplot also illustrates a median score that appears to be close to 0.78, indicating acceptable performance, with numerous low-performing outliers, suggesting some challenges with achieving consistently high nDCG scores.\\
\begin{figure}[h]  
    \centering
    \includegraphics[width=1\linewidth]{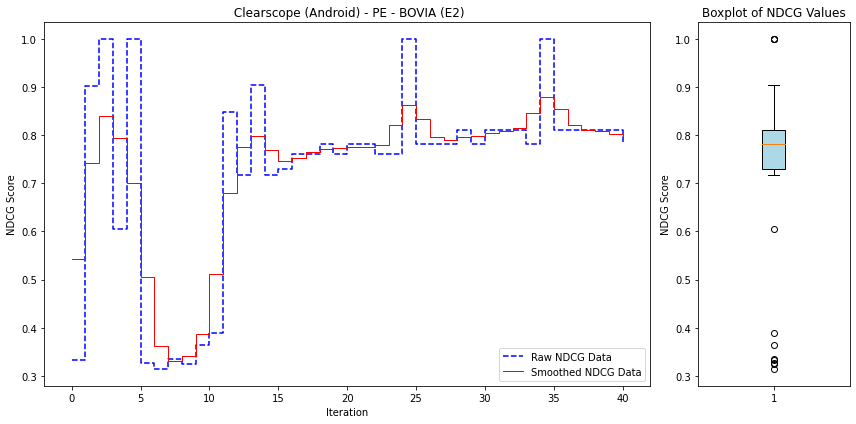}
    \caption{nDCG score variation over Active Learning iterations for the Android (Clearscope) PE dataset (Bovia E2 scenario).}
    \label{fig:androidPEBovia}
\end{figure}
Figures \ref{fig:AL_BSD_Pandex}, \ref{fig:AL_BSD_Bovia},  \ref{fig:AL_Linux_Pandex}, \ref{fig:AL_Linux_Bovia}, \ref{fig:AL_Windows_Pandex}, \ref{fig:AL_Windows_Bovia}, \ref{fig:AL_Android_Pandex}, and \ref{fig:AL_Android_Bovia} represent the nDCG variations for all remaining datasets under both attack scenarios. Each subplot shows both the raw and smoothed nDCG scores across 40 iterations, alongside a boxplot summarizing the distribution of nDCG values for each dataset. These figures cover the BSD, Linux, Windows and Android operating system and the five datasets: PA, PE, PP, PX, and PN. The smoothed curves provide a clearer view of the trends and fluctuations in ranking quality, while the boxplots offer insights into the overall distribution and outliers in the nDCG scores. The overall trend for the larger datasets in these figures is upward, showing consistent improvement in nDCG scores as iterations progress. While some datasets exhibit fluctuations due to data sparsity (e.g. Linux-PX/PP in Pandex scenario), the nDCG scores after applying active learning are generally higher than the initial values obtained in Table \ref{tab:winning-algo}, indicating that active learning has positively impacted the ranking performance.\\
Table \ref{ALADAEN-baseline-AL} summarizes the assessment of the performance of ALADAEN algorithm in both settings: (1) without active learning (baseline)—previously reported in Table \ref{ndcgscoresevalAll}— and (2) with active learning, across different operating systems (BSD, Windows, Linux, and Android) and datasets (PA, PE, PX, PP, and PN) under two attack scenarios (Pandex and Bovia). We report the maximum, mean, and median nDCG scores to provide a holistic assessment of the model's performance during active learning. The maximum nDCG score reflects the model's peak performance during active learning, indicating its best ability to rank anomalies. The mean provides an overall average, showcasing the model's consistency across iterations. The median, less affected by outliers, offers a more reliable measure of typical performance, especially in cases of fluctuating results. It's worth noting that during the active learning process, when the nDCG score reaches 1, indicating perfect ranking performance, the learning process could have been ideally stopped, as the model has already achieved the best possible outcome and further iterations are unlikely to yield significant improvements.\\
The main observations are as follows: 
\begin{itemize}
    \item BSD Operating System: Across almost all datasets in both attack scenarios (Pandex and Bovia), active learning consistently improved the nDCG scores compared to the baseline.
The maximum nDCG score with active learning often reached 1.0, especially in the Bovia attack scenario.
There were some fluctuations in the median and mean nDCG scores for smaller datasets (e.g., PN), likely due to data sparsity. However, active learning still resulted in higher nDCG scores overall, particularly in scenarios like PA and PE where improvements were most pronounced.

 \item Windows Operating System: Active learning showed significant improvements across most datasets and attack scenarios.
In the Pandex scenario, the maximum nDCG score consistently reached high values, with the mean and median values substantially higher than the baseline.
The Bovia scenario also saw notable improvements, particularly in datasets like PA, PX and PP, where active learning helped push the nDCG scores to their highest possible values.
Smaller datasets showed lower mean and median values, but the trend is upward with active learning.
 
 \item Linux Operating System: The application of active learning improved the performance in the majority of the datasets, with notable increases in the PA, PE, PX, and PN datasets.
In particular, the maximum nDCG scores were consistently higher in the active learning setting compared to the baseline.
Active learning reduced the variation between mean and median values, stabilizing the performance across multiple iterations.

 \item Android Operating System: Active learning resulted in higher maximum and median nDCG scores, especially in datasets like PA and PE.
Notable improvements were observed in both Pandex and Bovia scenarios, with active learning increasing the maximum nDCG scores to 1.0 in several cases.

\end{itemize}

\begin{figure} 
    \centering
    \includegraphics[width=1\linewidth]{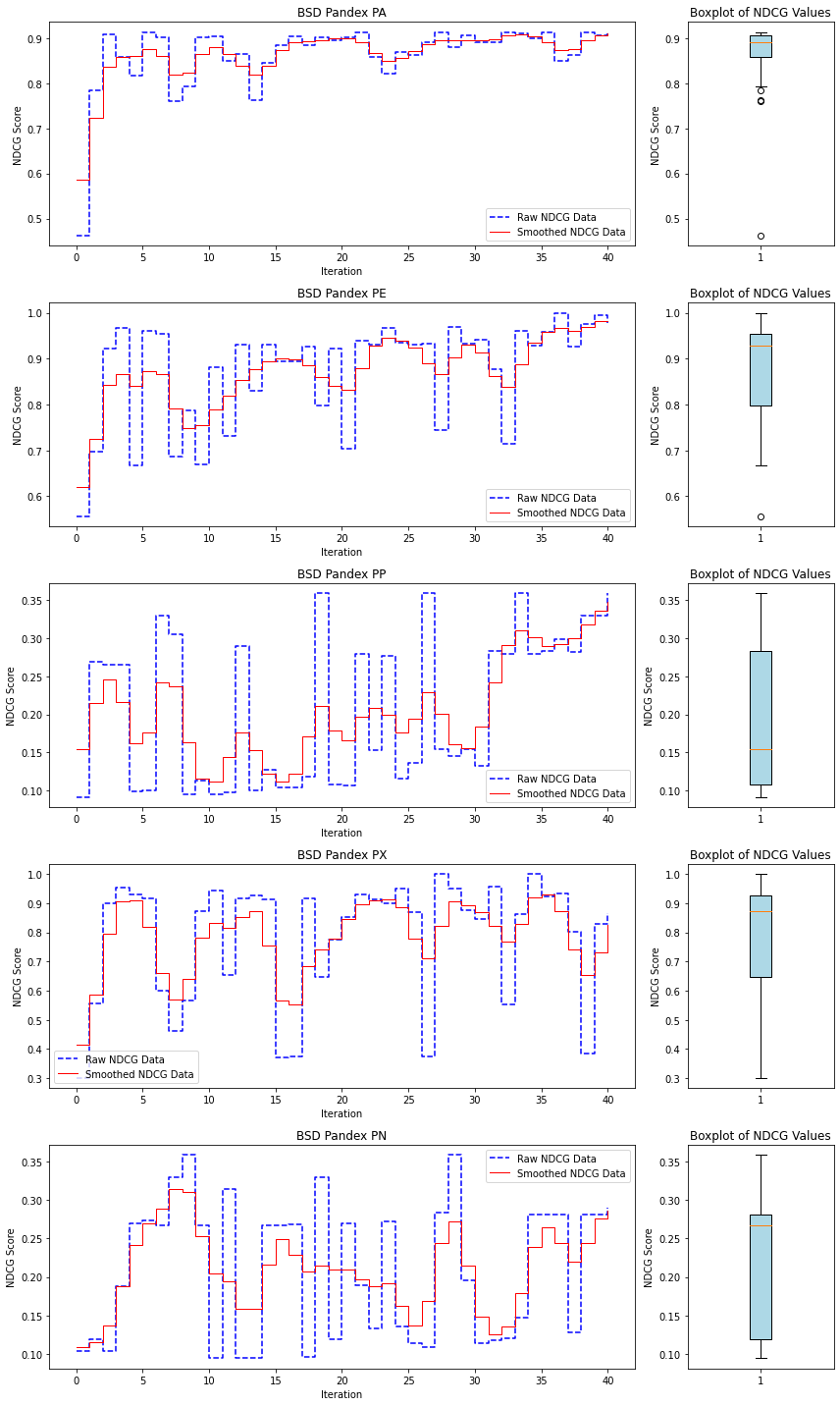}
    \caption{nDCG score variation over Active Learning iterations for BSD datasets (Pandex E1 scenario).}
    \label{fig:AL_BSD_Pandex}
\end{figure}
\begin{figure}
    \centering
    \includegraphics[width=1\linewidth]{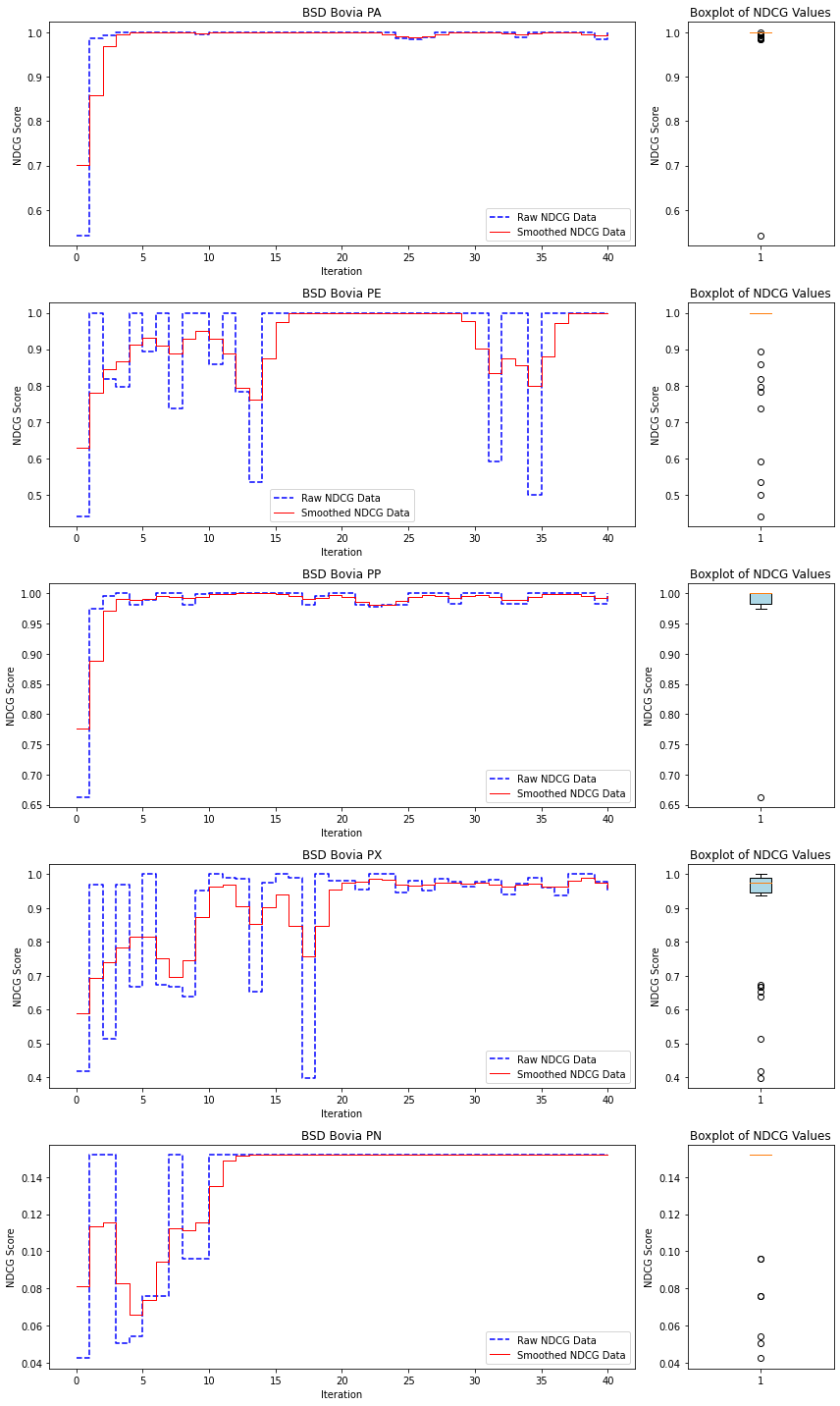}
    \caption{nDCG score variation over Active Learning iterations for BSD datasets (Bovia E2 scenario).}
    \label{fig:AL_BSD_Bovia}
\end{figure}

\begin{figure} 
    \centering
    \includegraphics[width=1\linewidth]{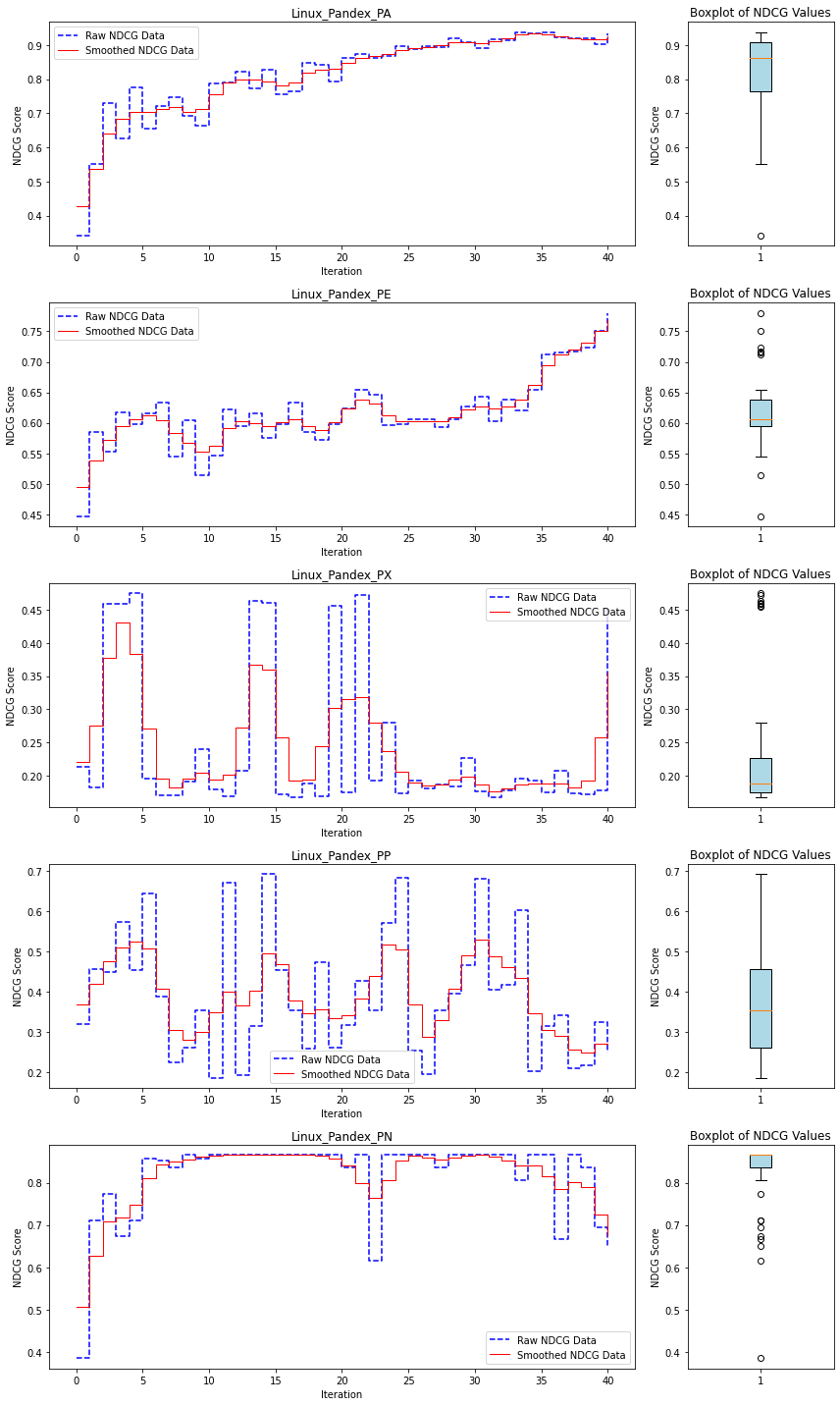}
    \caption{nDCG score variation over Active Learning iterations for Linux datasets (Pandex E1 scenario).}
    \label{fig:AL_Linux_Pandex}
\end{figure}
\begin{figure}
    \centering
    \includegraphics[width=1\linewidth]{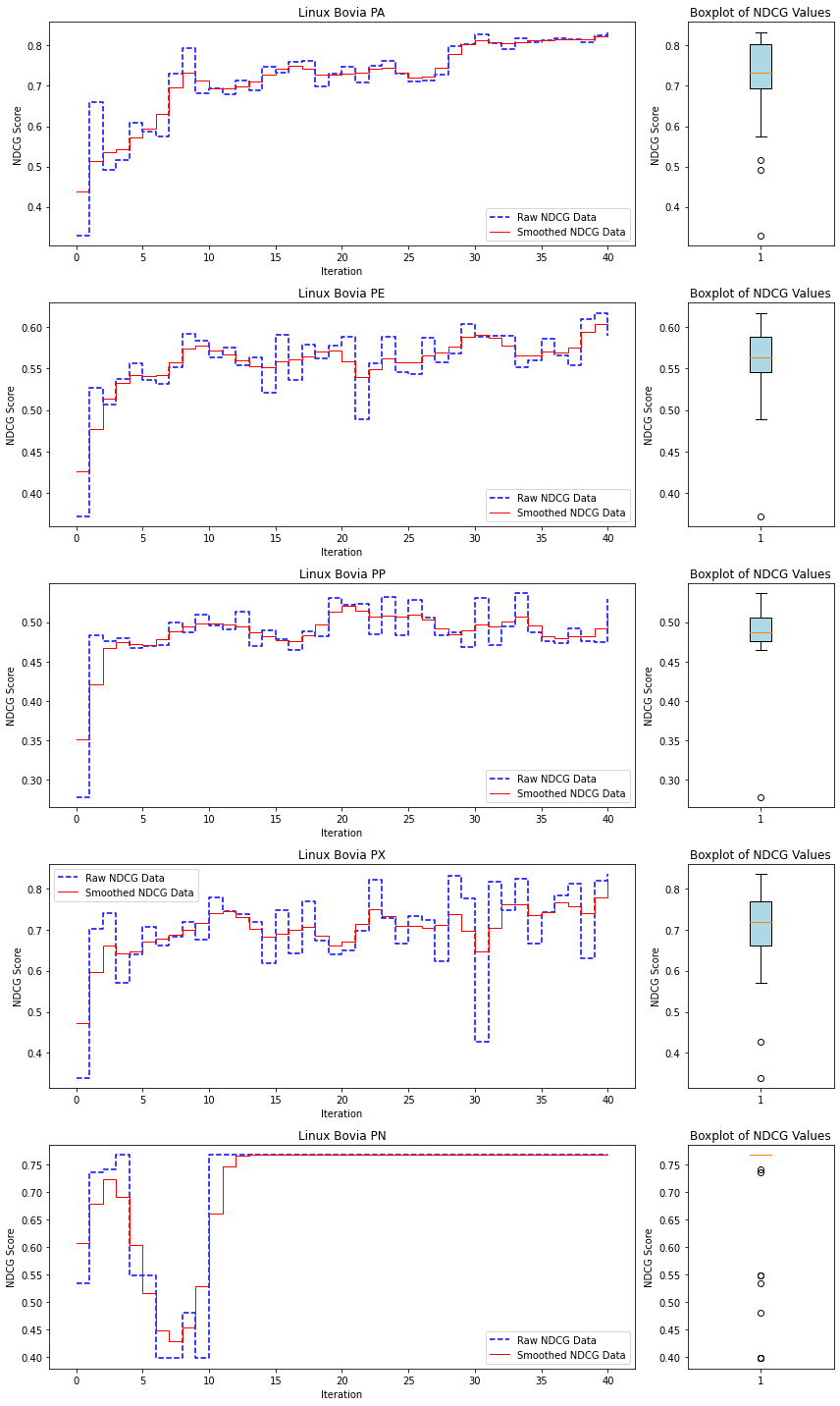}
    \caption{nDCG score variation over Active Learning iterations for Linux datasets (Bovia E2 scenario).}
    \label{fig:AL_Linux_Bovia}
\end{figure}
\begin{figure} 
    \centering
    \includegraphics[width=1\linewidth]{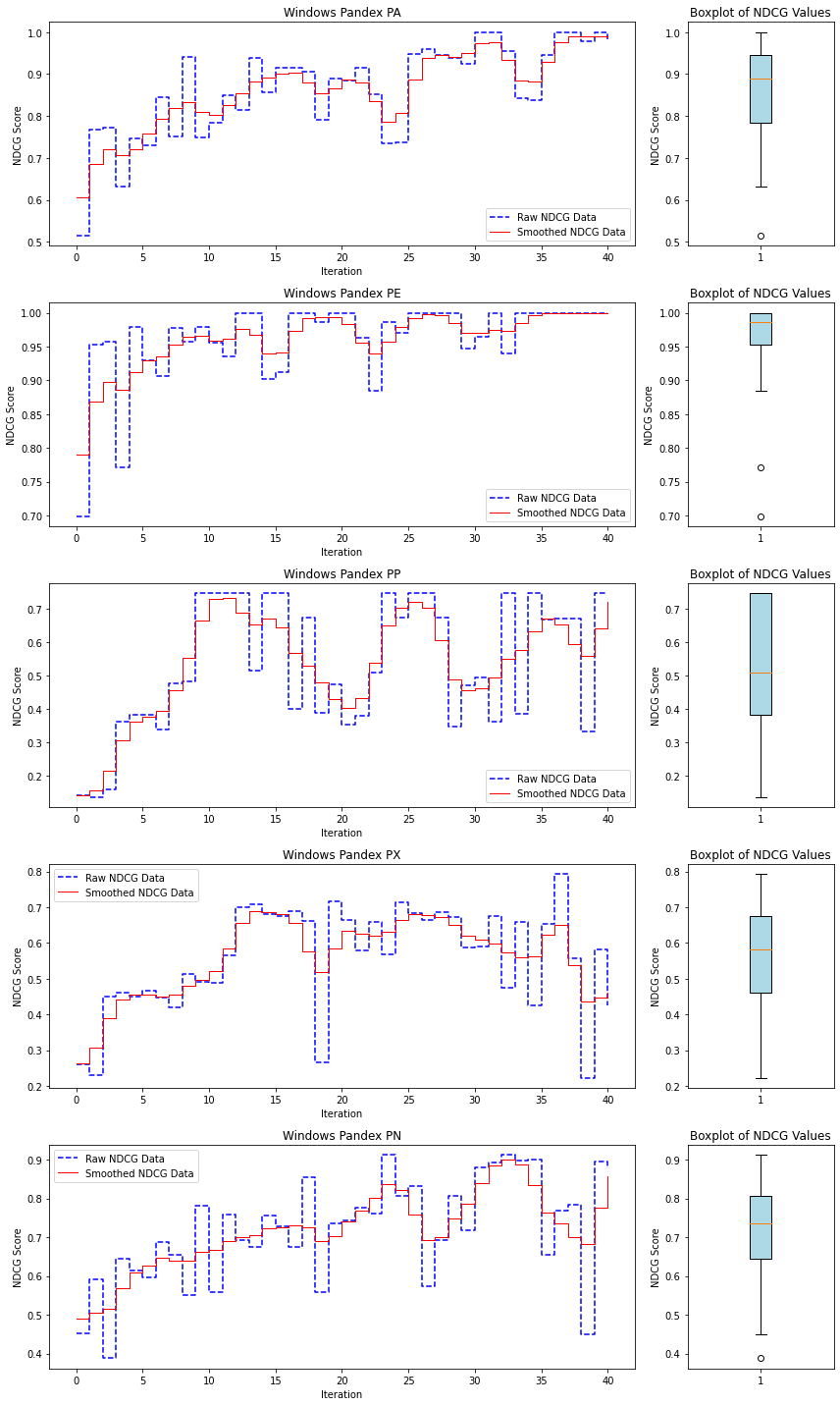}
    \caption{nDCG score variation over Active Learning iterations for Windows datasets (Pandex E1 scenario).}
    \label{fig:AL_Windows_Pandex}
\end{figure}
\begin{figure} 
    \centering
    \includegraphics[width=1\linewidth]{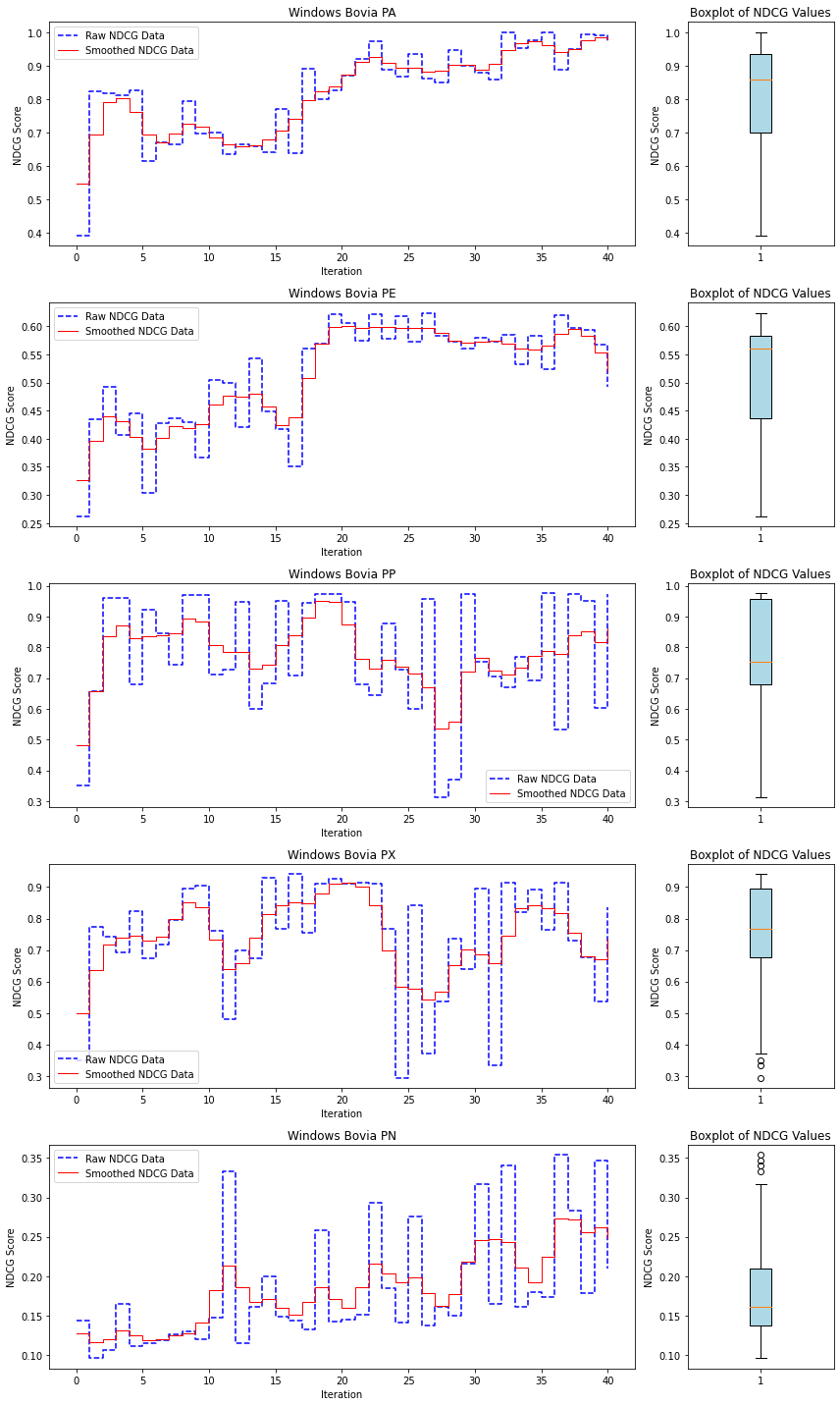}
    \caption{nDCG score variation over Active Learning iterations for Windows datasets (Bovia E2 scenario).}
    \label{fig:AL_Windows_Bovia}
\end{figure}
\begin{figure}  
    \centering
    \includegraphics[width=1\linewidth]{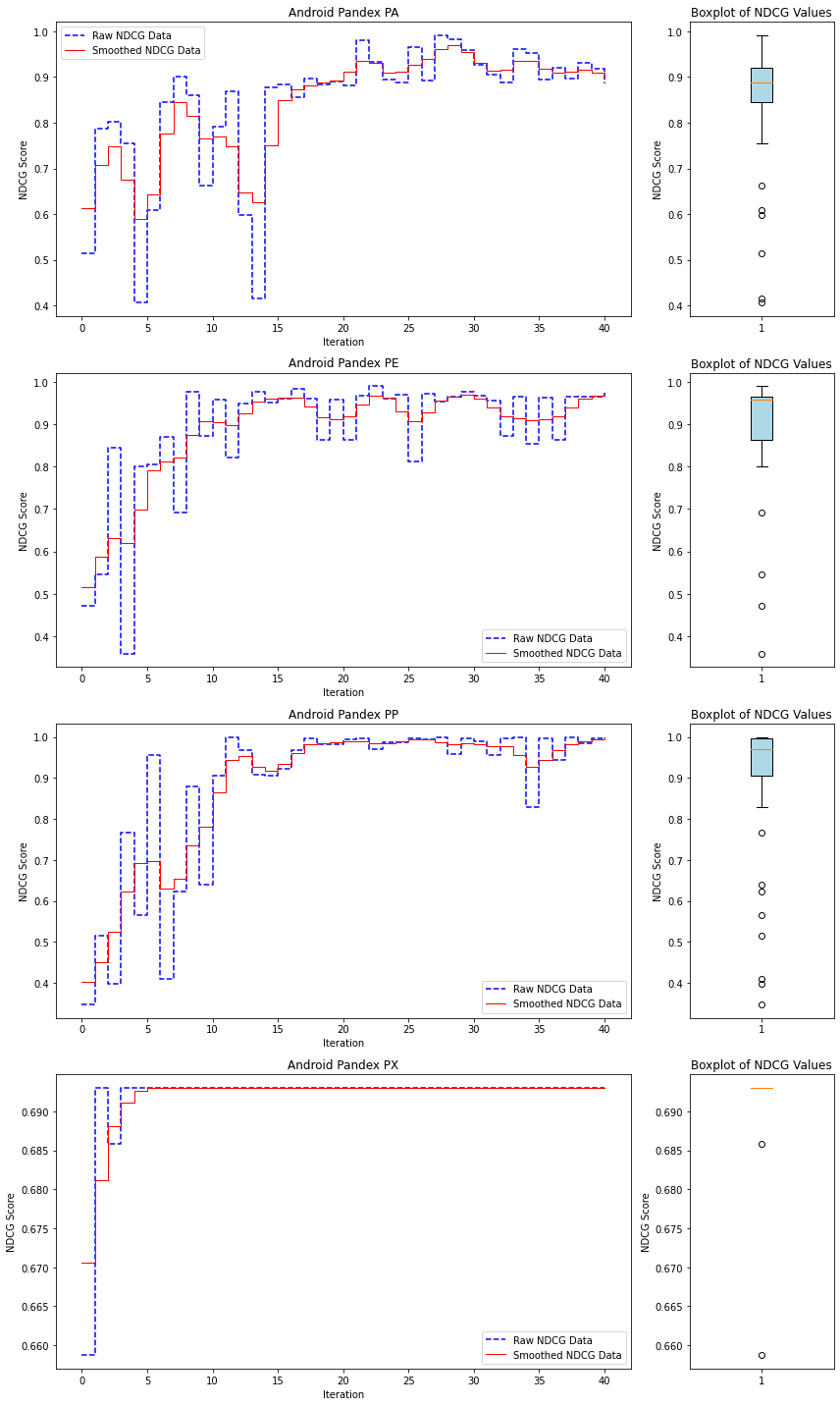}
    \caption{nDCG score variation over Active Learning iterations for Android datasets (Pandex E1 scenario).}
    \label{fig:AL_Android_Pandex}
\end{figure}
\begin{figure}
    \centering
    \includegraphics[width=1\linewidth]{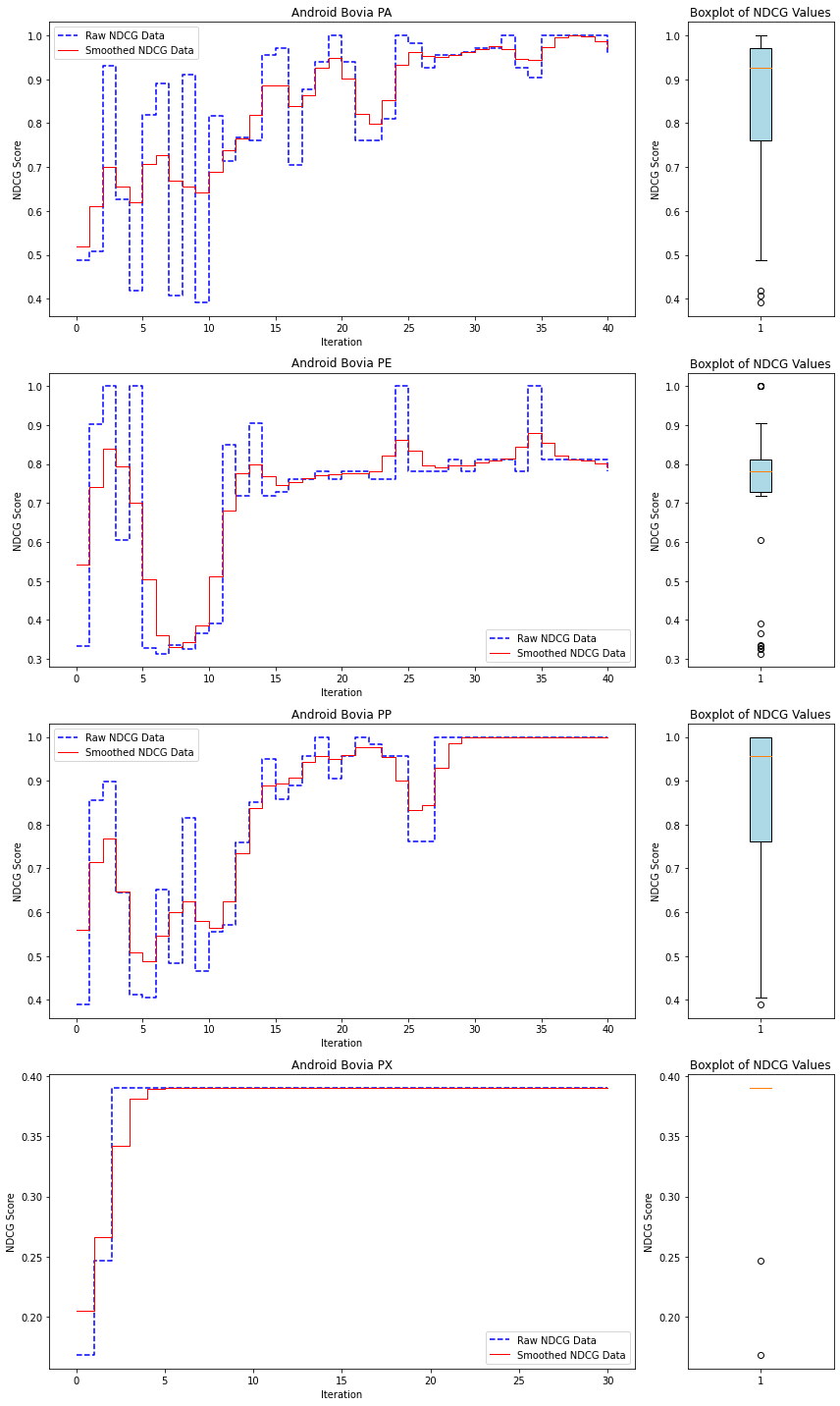}
    \caption{nDCG score variation over Active Learning iterations for Android datasets (Bovia E2 scenario).}
    \label{fig:AL_Android_Bovia}
\end{figure}
\begin{table*}

\centering
\Rotatebox{90}{
\scriptsize
\begin{tabular}{lc|cccc|cccc|cccc|cccc|cccc}
\hline

\multirow{3}{*}{\textbf{\rotatebox{90}{Operating System}}} & \multirow{3}{*}{\textbf{\rotatebox{90}{Attack Scenario}}} & \multicolumn{4}{c|}{\textbf{PA}}                                                                                             & \multicolumn{4}{c|}{\textbf{PE}}                                                                                                   & \multicolumn{4}{c|}{\textbf{PX}}                                                                                                   & \multicolumn{4}{c|}{\textbf{PP}}                                                                                                   & \multicolumn{4}{c}{\textbf{PN}}                                                                                               \\
\cline{3-22} 
                                           &                                           & \multirow{2}{*}{\textbf{\rotatebox{90}{Baseline}}} & \multicolumn{3}{c|}{\textbf{\rotatebox{90}{Active Learning}}}                                           & \textbf{\rotatebox{90}{Baseline}}    & \multicolumn{3}{c|}{\textbf{\rotatebox{90}{Active Learning}}}                                                               & \textbf{\rotatebox{90}{Baseline}}    & \multicolumn{3}{c|}{\textbf{\rotatebox{90}{Active Learning}}}                                                               & \textbf{\rotatebox{90}{Baseline}}    & \multicolumn{3}{c|}{\textbf{\rotatebox{90}{Active Learning}}}                                                               & \textbf{\rotatebox{90}{Baseline}}    & \multicolumn{3}{c}{\textbf{\rotatebox{90}{Active Learning}}}                                                               \\ \cline{4-6} \cline{8-10} \cline{12-14} \cline{16-18} \cline{20-22} 
                                           &                                           &                                    & \multicolumn{1}{c}{\textbf{\rotatebox{90}{Max}}} & \textbf{\rotatebox{90}{Mean}} & \multicolumn{1}{c|}{\textbf{\rotatebox{90}{Median}}} & \multicolumn{1}{l}{} & \multicolumn{1}{c}{\textbf{\rotatebox{90}{Max}}} & \multicolumn{1}{c}{\textbf{\rotatebox{90}{Mean}}} & \multicolumn{1}{c|}{\textbf{\rotatebox{90}{Median}}} & \multicolumn{1}{l}{} & \multicolumn{1}{c}{\textbf{\rotatebox{90}{Max}}} & \multicolumn{1}{c}{\textbf{\rotatebox{90}{Mean}}} & \multicolumn{1}{c|}{\textbf{\rotatebox{90}{Median}}} & \multicolumn{1}{l}{} & \multicolumn{1}{c}{\textbf{\rotatebox{90}{Max}}} & \multicolumn{1}{c}{\textbf{\rotatebox{90}{Mean}}} & \multicolumn{1}{c|}{\textbf{\rotatebox{90}{Median}}} & \multicolumn{1}{l}{} & \multicolumn{1}{c}{\textbf{\rotatebox{90}{Max}}} & \multicolumn{1}{c}{\textbf{\rotatebox{90}{Mean}}} & \multicolumn{1}{c}{\textbf{\rotatebox{90}{Median}}} \\ 
                                           \hline
BSD                              & Pandex                                    & 0.88                                                  & 0.91                                            & 0.87                                             & 0.89                                               & 0.75                                                 & 1                                               & 0.87                                             & 0.93                                               & 0.78                                                 & 1                                               & 0.78                                             & 0.87                                               & 0.65                                                 & 0.36                                            & 0.21                                             & 0.15                                               & 0.35                                                 & 0.36                                            & 0.21                                             & 0.27                                               \\ \cline{2-22} 
                                           & Bovia                                     & 0.79                                                                  & 1                                               & 0.98                                             & 1                                                  & 0.67                                                 & 1                                               & 0.93                                             & 1                                                  & 0.77                                                 & 1                                               & 0.90                                             & 0.97                                               & 0.98                                      & 1                                               & 0.99                                             & 1                                                  & 0.15                                                 & 0.15                                            & 0.14                                             & 0.15                                               \\ \hline
Windows                             & Pandex                                    & 0.72                                                                  & 1                                               & 0.86                                             & 0.89                                               & 0.82                                        & 1                                               & 0.96                                             & 0.97                                               & 0.24                                                 & 0.79                                            & 0.56                                             & 0.58                                               & 0.17                                                 & 0.75                                            & 0.53                                             & 0.51                                               & 0.65                                                 & 0.91                                            & 0.72                                             & 0.73                                               \\ \cline{2-22} 
                                           & Bovia                                     & 0.45                                                         & 1                                               & 0.82                                             & 0.86                                               & 0.31                                                 & 0.62                                            & 0.51                                             & 0.56                                               & 0.43                                                 & 0.94                                            & 0.72                                             & 0.77                                               & 0.43                                                 & 0.97                                            & 0.78                                             & 0.75                                               & 0.4                                                  & 0.36                                            & 0.16                                             & 0.19                                               \\ \hline
Linux                            & Pandex                                    & 0.77                                                     & 0.94                                            & 0.82                                             & 0.84                                               & 0.65                                                 & 0.78                                            & 0.62                                             & 0.61                                               & 0.36                                                 & 0.69                                            & 0.39                                             & 0.36                                               & 0.23                                                 & 0.47                                            & 0.24                                             & 0.19                                               & 0.45                                                 & 0.87                                            & 0.82                                             & 0.87                                               \\ \cline{2-22} 
                                           & Bovia                                     & 0.87                                                        & 0.84                                            & 0.72                                             & 0.73                                               & 0.56                                                 & 0.62                                            & 0.56                                             & 0.56                                               & 0.38                                                 & 0.84                                            & 0.70                                             & 0.72                                               & 0.47                                                 & 0.54                                            & 0.49                                             & 0.48                                               & 0.77                                                 & 0.77                                            & 0.72                                             & 0.77                                               \\ \hline
Android                      & Pandex                                    & 0.80                                                                  & 0.99                                            & 0.84                                             & 0.89                                               & 0.87                                   & 0.99                                            & 0.87                                             & 0.96                                               & 0.45                                                 & 1                                               & 0.88                                             & 0.97                                               & NA                                                   & NA                                              & NA                                               & NA                                                 & 0.68                                                 & 0.69                                            & 0.69                                             & 0.69                                               \\ \cline{2-22} 
                                           & Bovia                                     & 0.52                                                                  & 1                                               & 0.85                                             & 0.93                                               & 0.67                                                 & 1                                               & 0.74                                             & 0.78                                               & 0.51                                                 & 1                                               & 0.85                                             & 0.96                                               & NA                                                   & NA                                              & NA                                               & NA                                                 & 0.41                                                 & 0.39                                            & 0.38                                             & 0.39                                               \\ \hline

\end{tabular}
}
\caption{Comparison of ALADAEN Performance with and without Active Learning (Baseline). For each dataset, the first column represents the scores where the algorithm is trained on the entire dataset from the start without any iterative data augmentation. The remaining columns (Max, Mean, Median) represent the scores when running Active Learning. The maximum captures the peak performance, the mean reflects the overall average behavior across all iterations, and the median offers a robust measure of typical performance by minimizing the impact of outliers. 
}
\label{ALADAEN-baseline-AL}
\end{table*}
%
%
%
%
%
%
\subsection{Analysis of Active Learning Gains}
\label{sec:active-learning-analysis}

\begin{figure}[t]
    \centering
    \includegraphics[width=\linewidth]{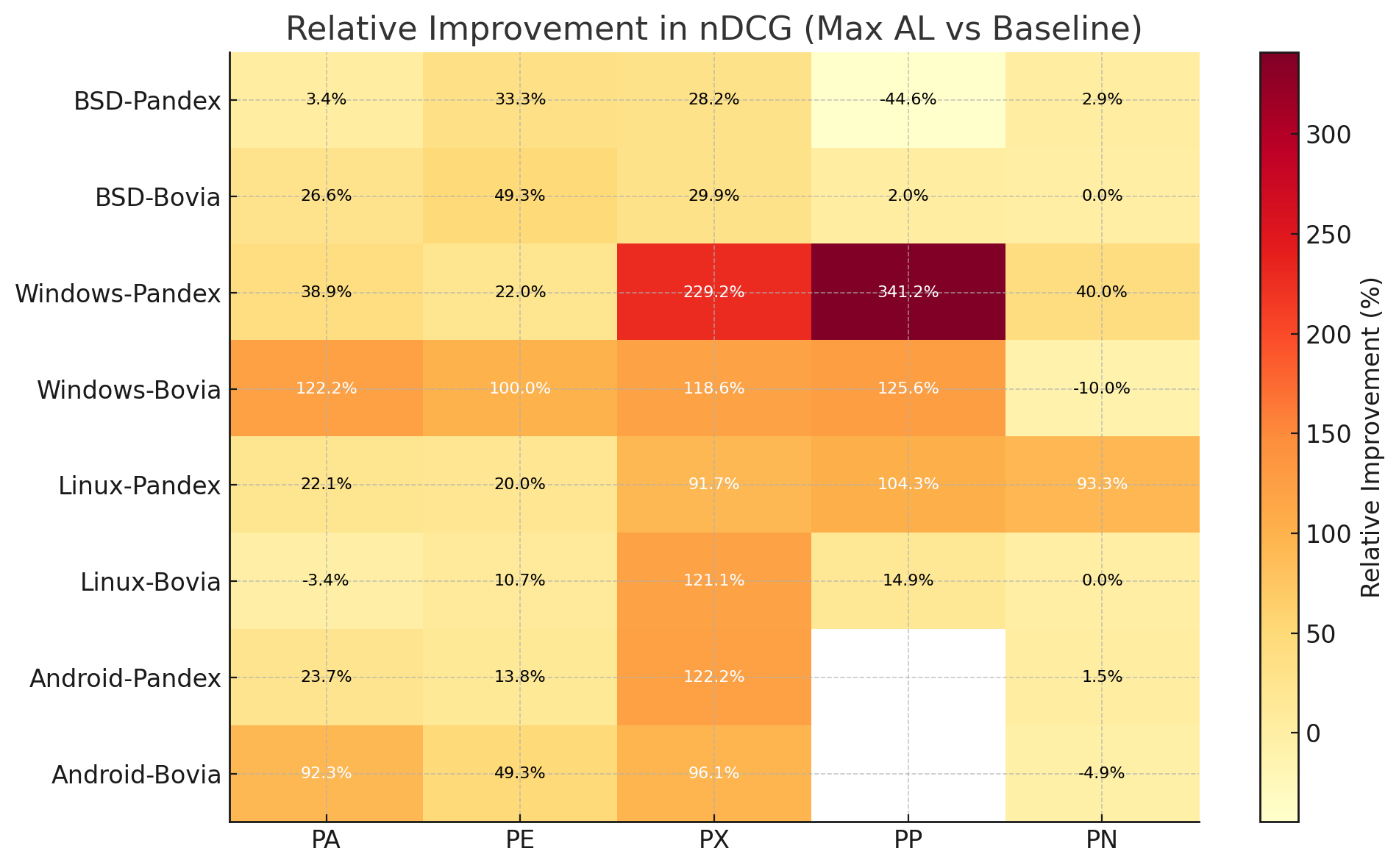}
    \caption{Relative improvement in nDCG (\%) for ALADAEN with active learning compared to the baseline 
    (active learning ablated). Values are computed as 
    $\Delta = (\mathrm{Max}_{AL} - \mathrm{Baseline}) / \mathrm{Baseline} \times 100$ 
    for each OS $\times$ scenario $\times$ view combination. 
    Deeper red indicates a larger relative gain. 
    Improvements are consistently positive, with the most pronounced gains observed for PX and PP views 
    (often exceeding $+100\%$), showing that active learning can more than double performance in the most 
    challenging configurations.}
    \label{fig:heatmap-improvement}
\end{figure}

Table~\ref{ALADAEN-baseline-AL} compares ALADAEN trained once without active learning (Baseline) 
against its performance during iterative oracle querying (Active Learning). 
The Baseline represents an ablation of the active learning component — 
no querying, no GAN augmentation , only a single training pass on the training database. 
The three remaining columns (Max, Mean, Median) summarize performance during active learning, 
capturing respectively the peak, average, and robust typical values across query iterations.\\
Figure~\ref{fig:heatmap-improvement} visualizes the {maximum relative improvement in nDCG} 
achieved during active learning across all OS $\times$ scenario $\times$ view combinations. 
The relative gain is computed as 
$\Delta = (\mathrm{Max}_{AL} - \mathrm{Baseline}) / \mathrm{Baseline} \times 100$. 
Because the maximum iteration is reported, all values are positive or zero, 
and the figure highlights the true potential of ALADAEN when its querying process reaches its best iteration. 
The strongest relative gains are observed for PX and PP views, where baseline performance is weakest. 
In several cases, the improvement exceeds $+100\%$, particularly for Windows and BSD scenarios, 
showing that active learning can more than double detection quality in the most challenging views.\\
In the previously reported nDCG curves, the figures (5-20) also show the convergence behavior of nDCG as a function of the number of queried samples, 
revealing that most gains occur in the first few iterations, confirming the label efficiency of the approach. 
The boxplots further summarize the distribution of relative improvements across all datasets, 
showing that gains are consistent across configurations and not driven by a small set of outliers.\\
Together, these results confirm that active learning not only consistently improves anomaly ranking performance 
but can achieve substantial peak improvements when allowed to query a small number of the most informative samples.

\subsection{\textbf{Runtime and Efficiency}}
In addition to accuracy, we evaluated the computational efficiency of ALADAEN. 
Across all DARPA datasets and views, inference and ranking 
(including reconstruction, score computation, sorting, and active learning selection) 
completed in $12.1 \pm 1.9$ minutes on average on a single Apple M1 Max CPU core 
with 64\,GB of RAM. 
This represents a substantial practical advantage: baselines such as FPOF, OC3, 
and OD frequently failed to terminate within $> 4$ hours on high-dimensional  views, underscoring their unsuitability for time-sensitive applications.\\
Table~\ref{tab:runtime} compares the average inference time across all methods. 
Among recent SOTA baselines, RT-APT and MAE-AD demonstrate efficient execution with average runtimes of $10.15$ and $11.4$ minutes respectively, confirming their suitability for real-time or near-real-time analysis. 
ALADAEN maintains comparable efficiency at $12.1 \pm 1.9$ minutes despite its dual-adversarial and active learning components, highlighting its optimized implementation and balanced computational cost. 
In contrast, traditional baselines such as VR-ARM and AVF exhibit moderate runtimes ($15$–$22$ minutes), while classical rule-mining approaches like FPOF and OC3 experience significant slowdowns, occasionally failing to complete within practical limits. 
Deep learning architectures with higher graph complexity, such as PGAE-APT and DeepOCCL, display longer runtimes ($>30$ minutes), underscoring their limited scalability for large-scale provenance data. 
Overall, ALADAEN achieves an ideal trade-off between accuracy and efficiency, confirming its operational readiness for deployment in Security Operations Center (SOC) environments where rapid anomaly prioritization is critical.
\\
The runtime of ALADAEN scales linearly with the number of samples and can be 
further reduced by mini-batching or GPU acceleration. 
Because ALADAEN is trained and reused for streaming inference, 
its per-process latency (below $10$\,ms per process) satisfies near-real-time 
requirements typical in Security Operations Center (SOC) workflows, 
where timely anomaly prioritization is critical for effective incident response.

\begin{table}[!ht]
\centering
\caption{Average inference + ranking runtime (minutes) across all datasets and views.
Values are computed over successful runs; FPOF, OC3, and OD exhibited DNFs 
on a subset of high-dimensional configurations ($>4$ h).}
\begin{tabular}{lcc}
\hline
\textbf{Method} & \textbf{Average Runtime (min)} & \textbf{Std. Dev. (min)} \\
\hline
RT-APT & $10.15$ & $3.1$ \\
MAE-AD& $11.4$ & $2.9$ \\
\textbf{ALADAEN (ours)} & $\mathbf{12.1}$ & $1.9$ \\
HD-APT & $12.54$ & $2.1$ \\
VR-ARM & $15.6$ & $1.0$ \\
AVF & $22.4$ & $1.2$ \\
FPOF & $26.0$ & $2.5$ (partial DNF) \\
OC3 & $27.0$ & $3.1$ (partial DNF) \\
PAGE-APT & $30.5$ & $2.6$ \\
OD  & $31.3$ & $4.0$ (partial DNF) \\
DeepOCCL  & $35.7$ & $3.2$ (partial DNF) \\
\hline
\end{tabular}
\label{tab:runtime}
\end{table}
\subsection{Discussions}
The integration of synthetic data generated by GAN plays a crucial role in enhancing the training process, especially when labeled data is limited. By mimicking the distribution of normal data, the GAN expands the training set, helping the model better capture normal behavior patterns. This additional data refines and stabilizes the decision boundary between normal and anomalous instances, leading to a clearer separation.\\
A major challenge with small data sets is overfitting, where the model struggles to generalize to unseen data. The variability introduced by the synthetic data acts as a regularizer, reducing overfitting and improving the model's robustness. By learning from both real and synthetic data, the model becomes more adept at generalizing to new data points.\\
Anomaly detection tasks often face a class imbalance, with normal instances far outnumbering anomalies. The GAN-generated data balances this disparity, allowing the AutoEncoder to learn the normal distribution more thoroughly. This improves the model's ability to detect subtle outliers and enhances its overall anomaly detection performance.\\
In the active learning process, synthetic data helps refine the decision boundary. The most uncertain data points, often near the decision boundary, are selected for labeling. By generating synthetic data around these points, the boundary is iteratively adjusted, improving the model’s ability to classify ambiguous instances with greater confidence. This iterative process, reinforced by both real and synthetic data, leads to a more accurate generalization.\\
The diversity provided by the GAN-generated data broadens the model’s understanding of normal behavior, reducing the likelihood of false positives. As a result, the decision boundary is more precisely defined, minimizing the misclassification of normal data and creating a sharper distinction between normal and anomalous data.

%
 %
%
 %
  %
  %
 %
  %
   %

\section{Conclusion}
In this paper, we presented \textit{ALADAEN}, an innovative framework  which combines Deep Neural Networks with Active Learning to address the complex problem of anomaly detection in environments with limited labeled data. The inclusion of a GAN-based data augmentation mechanism further strengthens the approach by effectively dealing with data scarcity, enabling the model to generate synthetic data that enriches the training process and enhances overall performance. This novel integration of GANs is particularly valuable in addressing class imbalance and improving generalization, making ALADAEN more adaptable than traditional anomaly detection methods.

The emphasis on active learning is well-aligned with the practical challenges of real-world applications, where acquiring labeled data is both costly and time-consuming. By selectively querying uncertain data points, the framework optimizes resource usage and gradually improves the model’s performance with minimal human intervention. This incremental approach to labeling, combined with data augmentation, ensures that even with a small amount of labeled data, the model can still refine its ability to detect anomalies effectively.

Additionally, the use of real-world provenance data from various operating systems (OS) strengthens the practical applicability of the framework. In scenarios like Advanced Persistent Threat (APT) detection, where anomalies are rare but critical, this approach is highly relevant. The framework’s ability to detect such anomalies across different OS environments further highlights its robustness and versatility in real-world security contexts.

Finally, the comprehensive performance evaluation, which compares \textit{ALADAEN} with other state-of-the-art anomaly detection methods, validates its effectiveness. The analysis clearly demonstrates the framework’s capacity to outperform traditional approaches, showcasing its ability to handle complex datasets and detect anomalies with higher accuracy. Additionally, since the baseline model was evaluated independently and compared against several state-of-the-art approaches without incorporating the active learning loop, it can also be interpreted as a form of ablation study highlighting the added value of active refinement.
\section{Future Work}
For future work, several promising directions can be explored to further enhance the capabilities and effectiveness of the ALADAEN framework.
For instance, we could focus on applying ALADAEN to more sophisticated and dynamic anomaly types, such as those emerging in IoT environments or evolving cyber-attacks. Expanding the framework to handle such complex, multi-modal anomalies would increase its applicability to a wider range of security threats.\\
We would also like to explore optimization of Active Learning strategies. While the current active learning approach relies on uncertainty-based sampling, exploring more advanced active learning techniques—such as diversity sampling or reinforcement learning-based strategies—could improve the efficiency of data selection. This may lead to faster convergence and more robust model performance with even fewer labeled samples.\\
We would also like to explore transfer learning and domain adaptation for investigating how ALADAEN can be adapted across different domains and datasets without retraining from scratch which could significantly reduce computational costs. Incorporating transfer learning methods could enable the model to generalize better when applied to new operating systems or attack scenarios.\\
Future work could explore dynamic GAN adaptation for improving the GAN-based data augmentation mechanism, making it more adaptive to the evolving distribution of data during the active learning process. Introducing dynamic feedback loops between the GAN and the active learning process could generate synthetic data that is more closely aligned with the most uncertain and informative samples. \\
\section*{Statements and Declarations}
\subsection*{Funding}
 This work was partly funded by the European Research Council (ERC)., ERC Skye 682315.
\subsection*{Competing Interests}
The authors declare that they have no known competing financial interests or personal relationships that could have appeared to influence the work reported in this paper.%
\section*{Author contributions}
SB: Conceptualization, Methodology, Software, Visualization. SB, JC: Data curation. SB, TR, JC: Investigation. SB, TR, JC: Writing.
\bibliography{references}

\end{document}